\RequirePackage{fix-cm}
\documentclass[smallextended]{svjour3}       
\smartqed  
\usepackage{graphicx}
\usepackage{bm}
\usepackage[svgnames,dvipsnames,table]{xcolor}
\usepackage{hyperref}
\usepackage{amsmath}
\usepackage{amssymb}
\usepackage[numbers]{natbib}
\usepackage{adjustbox}
\usepackage{multirow}
\usepackage{tikz}
\usepackage{pgfplots}
\usepgfplotslibrary{patchplots}
\usetikzlibrary{trees,shapes,snakes,calc,matrix,positioning,fit,arrows,patterns,backgrounds,quotes,arrows.meta}
\usepackage{siunitx}
\usepackage{multicol}
\usepackage{caption}
\usepackage{subcaption} 
\usetikzlibrary{shapes.geometric}
\usetikzlibrary{shapes.arrows}
\usepgfplotslibrary{fillbetween}
\usepackage{array}
\usepackage{tabularx}
\usepackage{lipsum} 
 \usepackage{collcell}
\usepackage{hhline}
\usetikzlibrary{spy, calc}
\usepackage{array}
\usepackage{tabularx}
\usepackage{xspace}

\usepackage{bbding}
\usepackage{pifont}
\usepackage{diagbox}
\usepackage{booktabs}
\usepackage{makecell}
\usetikzlibrary{fit}
\usepackage{rotating}

\newcolumntype{P}[1]{>{\centering\arraybackslash}m{#1}}
\newcolumntype{R}[1]{>{\raggedright\arraybackslash}m{#1}}%

\tikzset{
  fitting node/.style={
    inner sep=0pt,
    fill=none,
    draw=none,
    reset transform,
    fit={(\pgf@pathminx,\pgf@pathminy) (\pgf@pathmaxx,\pgf@pathmaxy)}
  },
  reset transform/.code={\pgftransformreset}
}

\newenvironment{customlegend}[1][]{%
        \begingroup
        \csname pgfplots@init@cleared@structures\endcsname
        \pgfplotsset{#1}%
    }{%
        \csname pgfplots@createlegend\endcsname
        \endgroup
    }%

\def\addlegendimage{\csname pgfplots@addlegendimage\endcsname}

\pgfplotsset{
  /pgfplots/xlabel near ticks/.style={
     /pgfplots/every axis x label/.style={
        at={(ticklabel cs:0.5)},anchor=near ticklabel
     }
  },
  /pgfplots/ylabel near ticks/.style={
     /pgfplots/every axis y label/.style={
        at={(ticklabel cs:0.5)},rotate=90,anchor=near ticklabel}
     }
  }

\newif\ifblackandwhitecycle
\gdef\patternnumber{0}

\pgfkeys{/tikz/.cd,
    zoombox paths/.style={
        draw=orange,
        very thick
    },
    black and white/.is choice,
    black and white/.default=static,
    black and white/static/.style={ 
        draw=white,   
        zoombox paths/.append style={
            draw=white,
            postaction={
                draw=black,
                loosely dashed
            }
        }
    },
    black and white/static/.code={
        \gdef\patternnumber{1}
    },
    black and white/cycle/.code={
        \blackandwhitecycletrue
        \gdef\patternnumber{1}
    },
    black and white pattern/.is choice,
    black and white pattern/0/.style={},
    black and white pattern/1/.style={    
            draw=white,
            postaction={
                draw=black,
                dash pattern=on 2pt off 2pt
            }
    },
    black and white pattern/2/.style={    
            draw=white,
            postaction={
                draw=black,
                dash pattern=on 4pt off 4pt
            }
    },
    black and white pattern/3/.style={    
            draw=white,
            postaction={
                draw=black,
                dash pattern=on 4pt off 4pt on 1pt off 4pt
            }
    },
    black and white pattern/4/.style={    
            draw=white,
            postaction={
                draw=black,
                dash pattern=on 4pt off 2pt on 2 pt off 2pt on 2 pt off 2pt
            }
    },
    zoomboxarray inner gap/.initial=5pt,
    zoomboxarray columns/.initial=2,
    zoomboxarray rows/.initial=2,
    subfigurename/.initial={},
    figurename/.initial={zoombox},
    zoomboxarray/.style={
        execute at begin picture={
            \begin{scope}[
                spy using outlines={%
                    zoombox paths,
                    width=0.5*\imagewidth / \pgfkeysvalueof{/tikz/zoomboxarray columns} - (\pgfkeysvalueof{/tikz/zoomboxarray columns} - 1) / \pgfkeysvalueof{/tikz/zoomboxarray columns} * \pgfkeysvalueof{/tikz/zoomboxarray inner gap} -\pgflinewidth,
                    height=\imageheight / \pgfkeysvalueof{/tikz/zoomboxarray rows} - (\pgfkeysvalueof{/tikz/zoomboxarray rows} - 1) / \pgfkeysvalueof{/tikz/zoomboxarray rows} * \pgfkeysvalueof{/tikz/zoomboxarray inner gap}-\pgflinewidth,
                    magnification=3,
                    every spy on node/.style={
                        zoombox paths
                    },
                    every spy in node/.style={
                        zoombox paths
                    }
                }
            ]
        },
        execute at end picture={
            \end{scope}
               \node at (image.north) [anchor=north,inner sep=0pt]{ \phantomimage};
               \node at (zoomboxes container.north) [anchor=north,inner sep=0pt]{ \phantomimage};
     \gdef\patternnumber{0}
        },
        spymargin/.initial=0.5em,
        zoomboxes xshift/.initial=1,
        zoomboxes right/.code=\pgfkeys{/tikz/zoomboxes xshift=1},
        zoomboxes left/.code=\pgfkeys{/tikz/zoomboxes xshift=-1},
        zoomboxes yshift/.initial=0,
        zoomboxes above/.code={
            \pgfkeys{/tikz/zoomboxes yshift=1},
            \pgfkeys{/tikz/zoomboxes xshift=0}
        },
        zoomboxes below/.code={
            \pgfkeys{/tikz/zoomboxes yshift=-1},
            \pgfkeys{/tikz/zoomboxes xshift=0}
        },
        caption margin/.initial=4ex,
    },
    adjust caption spacing/.code={},
    image container/.style={
        inner sep=0pt,
        at=(image.north),
        anchor=north,
        adjust caption spacing
    },
    zoomboxes container/.style={
        inner sep=0pt,
        at=(image.north),
        anchor=north,
        name=zoomboxes container,
        xshift=\pgfkeysvalueof{/tikz/zoomboxes xshift}*(\imagewidth+\pgfkeysvalueof{/tikz/spymargin}),
        yshift=\pgfkeysvalueof{/tikz/zoomboxes yshift}*(\imageheight+\pgfkeysvalueof{/tikz/spymargin}+\pgfkeysvalueof{/tikz/caption margin}),
        adjust caption spacing
    },
    calculate dimensions/.code={
        \pgfpointdiff{\pgfpointanchor{image}{south west} }{\pgfpointanchor{image}{north east} }
        \pgfgetlastxy{\imagewidth}{\imageheight}
        \global\let\imagewidth=\imagewidth
        \global\let\imageheight=\imageheight
        \gdef\columncount{1}
        \gdef\rowcount{1}
        
    },
    image node/.style={
        inner sep=0pt,
        name=image,
        anchor=south west,
        append after command={
            [calculate dimensions]
            node [image container,subfigurename=\pgfkeysvalueof{/tikz/figurename}-image] {\phantomimage}
            node [zoomboxes container,subfigurename=\pgfkeysvalueof{/tikz/figurename}-zoom] {\phantomimage}
        }
    },
    color code/.style={
        zoombox paths/.append style={draw=#1}
    },
    connect zoomboxes/.style={
    spy connection path={\draw[draw=none,zoombox paths] (tikzspyonnode) -- (tikzspyinnode);}
    },
    help grid code/.code={
        \begin{scope}[
                x={(image.south east)},
                y={(image.north west)},
                font=\footnotesize,
                help lines,
                overlay
            ]
            \foreach \x in {0,1,...,9} { 
                \draw(\x/10,0) -- (\x/10,1);
                \node [anchor=north] at (\x/10,0) {0.\x};
            }
            \foreach \y in {0,1,...,9} {
                \draw(0,\y/10) -- (1,\y/10);                        \node [anchor=east] at (0,\y/10) {0.\y};
            }
        \end{scope}    
    },
    help grid/.style={
        append after command={
            [help grid code]
        }
    },
}

\newcommand\phantomimage{%
    \phantom{%
        \rule{\imagewidth}{\imageheight}%
    }%
}
\newcommand\zoombox[2][]{
    \begin{scope}[zoombox paths]
        \pgfmathsetmacro\xpos{
            (\columncount-1)*(\imagewidth / \pgfkeysvalueof{/tikz/zoomboxarray columns} + \pgfkeysvalueof{/tikz/zoomboxarray inner gap} / \pgfkeysvalueof{/tikz/zoomboxarray columns} ) + \pgflinewidth
        }
        \pgfmathsetmacro\ypos{
            (\rowcount-1)*( \imageheight / \pgfkeysvalueof{/tikz/zoomboxarray rows} + \pgfkeysvalueof{/tikz/zoomboxarray inner gap} / \pgfkeysvalueof{/tikz/zoomboxarray rows} ) + 0.5*\pgflinewidth
        }
        \edef\dospy{\noexpand\spy [
            #1,
            zoombox paths/.append style={
                black and white pattern=\patternnumber
            },
            every spy on node/.append style={#1},
            x=\imagewidth,
            y=\imageheight
        ] on (#2) in node [anchor=north west] at ($(zoomboxes container.north west)+(\xpos pt,-\ypos pt)$);}
        \dospy
        \pgfmathtruncatemacro\pgfmathresult{ifthenelse(\columncount==\pgfkeysvalueof{/tikz/zoomboxarray columns},\rowcount+1,\rowcount)}
        \global\let\rowcount=\pgfmathresult
        \pgfmathtruncatemacro\pgfmathresult{ifthenelse(\columncount==\pgfkeysvalueof{/tikz/zoomboxarray columns},1,\columncount+1)}
        \global\let\columncount=\pgfmathresult
        \ifblackandwhitecycle
            \pgfmathtruncatemacro{\newpatternnumber}{\patternnumber+1}
            \global\edef\patternnumber{\newpatternnumber}
        \fi
    \end{scope}
}

\newcommand\magimage[5]{
\begin{tikzpicture}[spy using outlines={rectangle,red,magnification=4,size=#5}]
\node[anchor=north west] {\includegraphics[width=#2]{#1}};
\spy on (#2*#3,-#2*#4) in node at (#2-0.5*#5,-#2-0.5*#5);
\end{tikzpicture}
}

\newcommand\magimageblue[5]{
\begin{tikzpicture}[spy using outlines={rectangle,blue,magnification=4,size=#5}]
\node[anchor=north west] {\includegraphics[width=#2]{#1}};
\spy on (#2*#3,-#2*#4) in node at (#2-0.5*#5,-#2-0.5*#5);
\end{tikzpicture}
}

\newcommand{\pref}[1]{\begin{tikz}[baseline]\node[prefstyle,anchor=base]at(0,0){\tiny#1};\end{tikz}}

\newcommand{\pureref}[1]{\begin{tikz}\node[purerefstyle]at(0,0){#1};\end{tikz}}

\tikzstyle{prefstyle}=[circle, draw=red!50, fill=red!50,
        text centered, anchor=north, text=white,inner sep=0pt,text width=8pt]
        
\tikzstyle{purerefstyle}=[text centered, anchor=center, text=red,inner sep=4pt]        
        
\tikzset{mypin/.style n args={3}{dot,pin={[pin edge={-,red!50},pin distance = #1,inner sep=0pt,minimum width=10pt]#2:{\pref{\tiny#3}}}}}     
        
 \tikzset{refpin/.style n args={3}{pin={[pin edge={thick,<-,red,>=stealth,},pin distance = #1,inner sep=0pt,minimum width=10pt]#2:{\pureref{#3}}}}}     

\tikzstyle{dot}=[draw,circle,fill=red!50,minimum size=1mm,inner sep=0pt]

\newcommand\magimageref[5]{
\begin{tikzpicture}
\begin{scope}[spy using outlines={rectangle,red,magnification=4,size=#5}]
\node[anchor=north west] {\includegraphics[width=#2]{#1}};
\spy on (#2*#3,-#2*#4) in node (a) at (#2-0.5*#5,-#2-0.5*#5);
\end{scope}
\node[refpin={0.2cm}{90}{4-lp},xshift=13pt,yshift=4pt] at(a){};
\node[refpin={0.2cm}{90}{5-lp},xshift=3pt,yshift=-3pt] at(a){};
\node[refpin={0.2cm}{90}{6-lp},xshift=-8pt,yshift=-10pt] at(a){};

\end{tikzpicture}
}

\def\layersepsmall{4pt}

\tikzset{ 
    layer/.style={
    rectangle,fill=black!25,minimum width=140pt,inner sep=3pt, outer sep=0pt,minimum height=15pt,rounded corners=3pt
    }
 }
\tikzset{ 
    conv/.style={
    layer, fill=blue!30
    }
 }
 \tikzset{ 
    lrelu/.style={
    layer, fill=YellowGreen!40
    }
 }
 \tikzset{ 
    bn/.style={
    layer, fill=red!30
    }
 }
  \tikzset{ 
    dconv/.style={
    layer, fill=yellow!30
    }
 }
  \tikzset{ 
    relu/.style={
    layer, fill=YellowGreen!70
    }
 }
   \tikzset{ 
    set/.style={
    layer, fill=blue!15
    }
 }
   \tikzset{ 
    tanh/.style={
    layer, fill=pink!30
    }
 }
\tikzset{
	block1/.pic={
	\node[conv,draw](conv) {};
	\node[bn,draw,below=\layersepsmall of conv](bn) {};
	\node[lrelu,draw,below =\layersepsmall of bn](lrelu) {};	
	}
}

\tikzset{
	block2/.pic={
	\node[conv,draw](conv1) {};
	\node[bn,draw,below= \layersepsmall of conv1](bn1) {};
	\node[relu,draw,below =\layersepsmall of bn1](lrelu) {};	
	\node[conv,draw,below =\layersepsmall of lrelu](conv2) {};
	\node[bn,draw,below=\layersepsmall of conv2](bn2) {};
	}
}

\tikzset{
	block3/.pic={
	\node[dconv,draw](dconv) {};
	\node[bn,draw,below=\layersepsmall of dconv](bn) {};
	\node[relu,draw,below =\layersepsmall of bn](lrelu) {};	
	}
}

\tikzset{
	net/.pic={
	\node[layer,fill=gray,minimum width=140pt,inner sep=0,minimum height=20pt,rounded corners=2pt](l1){};
	\node[layer,fill=gray,minimum width=140pt,inner sep=0,minimum height=20pt,rounded corners=2pt,below= 3pt of l1](l2){};
	\node[layer,fill=gray,minimum width=140pt,inner sep=0,minimum height=20pt,rounded corners=2pt,below= 3pt of l2](l3){};
	}

}

\tikzset{
  annotated cuboid/.pic={
    \tikzset{%
      every edge quotes/.append style={midway, auto},
      /cuboid/.cd,
      #1
    }
    \draw [every edge/.append style={pic actions, densely dashed, opacity=.5}, pic actions]
    (0,0,0) coordinate (o) -- ++(-\cubescale*\cubex,0,0) coordinate (a) -- ++(0,-\cubescale*\cubey,0) coordinate (b) edge coordinate [pos=1] (g) ++(0,0,-\cubescale*\cubez)  -- ++(\cubescale*\cubex,0,0) coordinate (c) -- cycle
    (o) -- ++(0,0,-\cubescale*\cubez) coordinate (d) -- ++(0,-\cubescale*\cubey,0) coordinate (e) edge (g) -- (c) -- cycle
    (o) -- (a) -- ++(0,0,-\cubescale*\cubez) coordinate (f) edge (g) -- (d) -- cycle;
    ;
  },
  /cuboid/.search also={/tikz},
  /cuboid/.cd,
  width/.store in=\cubex,
  height/.store in=\cubey,
  depth/.store in=\cubez,
  units/.store in=\cubeunits,
  scale/.store in=\cubescale,
  width=10,
  height=10,
  depth=10,
  units=cm,
  scale=.1,
}

\newcommand{\parapp}[4]{%
\fill[#4,opacity=.5] (0,0,0)-- (#1,0,0) -- (#1,#3,0)  -- (0,#3,0) --cycle;
\fill[#4,opacity=.5] (0,0,#2)-- (#1,0,#2) -- (#1,#3,#2)  -- (0,#3,#2) --cycle;
\fill[#4,opacity=.5] (0,#3,0)-- (0,#3,#2) -- (#1,#3,#2) -- (#1,#3,0)--cycle;
\fill[#4,opacity=.5] (0,0,0)-- (0,0,#2) -- (#1,0,#2) -- (#1,0,0)--cycle; 
\draw[] (0,0,#2) -- (#1,0,#2) -- (#1,#3,#2) --(0,#3,#2) --(0,0,#2)
        (#1,0,#2) -- (#1,0,0)  -- (#1,#3,0) --(0,#3,0) -- (0,#3,#2)    
        (#1,#3,#2) -- (#1,#3,0);
\draw[dashed,opacity=0.4] (0,0,0) -- (0,0,#2) (0,0,0)-- (#1,0,0) (0,0,0)-- (0,#3,0);

}  

\newcommand{\parappdash}[4]{%
\fill[#4,opacity=.5] (0,0,0)-- (#1,0,0) -- (#1,#3,0)  -- (0,#3,0) --cycle;
\fill[#4,opacity=.5] (0,0,#2)-- (#1,0,#2) -- (#1,#3,#2)  -- (0,#3,#2) --cycle;
\fill[#4,opacity=.5] (0,#3,0)-- (0,#3,#2) -- (#1,#3,#2) -- (#1,#3,0)--cycle;
\fill[#4,opacity=.5] (0,0,0)-- (0,0,#2) -- (#1,0,#2) -- (#1,0,0)--cycle; 
\draw[dashed] (0,0,#2) --  (#1,0,#2) -- (#1,#3,#2) --(0,#3,#2) --(0,0,#2)
        (#1,0,#2) -- (#1,0,0)  -- (#1,#3,0) --(0,#3,0) -- (0,#3,#2)    
        (#1,#3,#2) -- (#1,#3,0);
\draw[] (#1,0,#2) -- (#1,#3,#2) --(0,#3,#2);
\draw[] (#1,0,#2) -- (#1,0,0)  -- (#1,#3,0) --(0,#3,0) -- (0,#3,#2)    
        (#1,#3,#2) -- (#1,#3,0);
\draw[dashed,opacity=0.4] (0,0,0) -- (0,0,#2) (0,0,0)-- (#1,0,0) (0,0,0)-- (0,#3,0);
}

\newif\ifcuboidshade
\newif\ifcuboidemphedge

\tikzset{
  cuboid/.is family,
  cuboid,
  shiftx/.initial=0,
  shifty/.initial=0,
  dimx/.initial=3,
  dimy/.initial=3,
  dimz/.initial=3,
  scale/.initial=1,
  densityx/.initial=1,
  densityy/.initial=1,
  densityz/.initial=1,
  rotation/.initial=0,
  anglex/.initial=0,
  angley/.initial=90,
  anglez/.initial=225,
  scalex/.initial=1,
  scaley/.initial=1,
  scalez/.initial=0.5,
  front/.style={draw=black,fill=white},
  top/.style={draw=black,fill=white},
  right/.style={draw=black,fill=white},
  shade/.is if=cuboidshade,
  shadecolordark/.initial=black,
  shadecolorlight/.initial=white,
  shadeopacity/.initial=0.15,
  shadesamples/.initial=16,
  emphedge/.is if=cuboidemphedge,
  emphstyle/.style={thick},
}

\newcommand{\tikzcuboidkey}[1]{\pgfkeysvalueof{/tikz/cuboid/#1}}

\newcommand{\tikzcuboid}[1]{
    \tikzset{cuboid,#1} 
  \pgfmathsetlengthmacro{\vectorxx}{\tikzcuboidkey{scalex}*cos(\tikzcuboidkey{anglex})*28.452756}
  \pgfmathsetlengthmacro{\vectorxy}{\tikzcuboidkey{scalex}*sin(\tikzcuboidkey{anglex})*28.452756}
  \pgfmathsetlengthmacro{\vectoryx}{\tikzcuboidkey{scaley}*cos(\tikzcuboidkey{angley})*28.452756}
  \pgfmathsetlengthmacro{\vectoryy}{\tikzcuboidkey{scaley}*sin(\tikzcuboidkey{angley})*28.452756}
  \pgfmathsetlengthmacro{\vectorzx}{\tikzcuboidkey{scalez}*cos(\tikzcuboidkey{anglez})*28.452756}
  \pgfmathsetlengthmacro{\vectorzy}{\tikzcuboidkey{scalez}*sin(\tikzcuboidkey{anglez})*28.452756}
  \begin{scope}[xshift=\tikzcuboidkey{shiftx}, yshift=\tikzcuboidkey{shifty}, scale=\tikzcuboidkey{scale}, rotate=\tikzcuboidkey{rotation}, x={(\vectorxx,\vectorxy)}, y={(\vectoryx,\vectoryy)}, z={(\vectorzx,\vectorzy)}]
    \pgfmathsetmacro{\steppingx}{1/\tikzcuboidkey{densityx}}
  \pgfmathsetmacro{\steppingy}{1/\tikzcuboidkey{densityy}}
  \pgfmathsetmacro{\steppingz}{1/\tikzcuboidkey{densityz}}
  \newcommand{\dimx}{\tikzcuboidkey{dimx}}
  \newcommand{\dimy}{\tikzcuboidkey{dimy}}
  \newcommand{\dimz}{\tikzcuboidkey{dimz}}
  \pgfmathsetmacro{\secondx}{2*\steppingx}
  \pgfmathsetmacro{\secondy}{2*\steppingy}
  \pgfmathsetmacro{\secondz}{2*\steppingz}
  \ifthenelse{\equal{\dimx}{1}}
    {\foreach \x in {\steppingx,...,\dimx}}
    {\foreach \x in {\steppingx,\secondx,...,\dimx}}
  {     \ifthenelse{\equal{\dimy}{1}}
    {\foreach \y in {\steppingy,...,\dimy}}
    {\foreach \y in {\steppingy,\secondy,...,\dimy}}
    {   \pgfmathsetmacro{\lowx}{(\x-\steppingx)}
      \pgfmathsetmacro{\lowy}{(\y-\steppingy)}
      \filldraw[cuboid/front] (\lowx,\lowy,\dimz) -- (\lowx,\y,\dimz) -- (\x,\y,\dimz) -- (\x,\lowy,\dimz) -- cycle;
    }
    }
    \ifthenelse{\equal{\dimx}{1}}
    {\foreach \x in {\steppingx,...,\dimx}}
    {\foreach \x in {\steppingx,\secondx,...,\dimx}}
  { \ifthenelse{\equal{\dimz}{1}}
    {\foreach \z in {\steppingz,...,\dimz}}
    {\foreach \z in {\steppingz,\secondz,...,\dimz}}
    {   \pgfmathsetmacro{\lowx}{(\x-\steppingx)}
      \pgfmathsetmacro{\lowz}{(\z-\steppingz)}
      \filldraw[cuboid/top] (\lowx,\dimy,\lowz) -- (\lowx,\dimy,\z) -- (\x,\dimy,\z) -- (\x,\dimy,\lowz) -- cycle;
        }
    }
    \ifthenelse{\equal{\dimy}{1}}
    {\foreach \y in {\steppingy,...,\dimy}}
    {\foreach \y in {\steppingy,\secondy,...,\dimy}}
  { \ifthenelse{\equal{\dimz}{1}}
    {\foreach \z in {\steppingz,...,\dimz}}
    {\foreach \z in {\steppingz,\secondz,...,\dimz}}
    {   \pgfmathsetmacro{\lowy}{(\y-\steppingy)}
      \pgfmathsetmacro{\lowz}{(\z-\steppingz)}
      \filldraw[cuboid/right] (\dimx,\lowy,\lowz) -- (\dimx,\lowy,\z) -- (\dimx,\y,\z) -- (\dimx,\y,\lowz) -- cycle;
    }
  }
  \ifcuboidemphedge
    \draw[cuboid/emphstyle] (0,\dimy,0) -- (\dimx,\dimy,0) -- (\dimx,\dimy,\dimz) -- (0,\dimy,\dimz) -- cycle;%
    \draw[cuboid/emphstyle] (0,\dimy,\dimz) -- (0,0,\dimz) -- (\dimx,0,\dimz) -- (\dimx,\dimy,\dimz);%
    \draw[cuboid/emphstyle] (\dimx,\dimy,0) -- (\dimx,0,0) -- (\dimx,0,\dimz);%
    \fi

  \end{scope}
}

\newcommand{\mycube}[1]{
   \tikzcuboid{%
    shiftx=0cm,%
    shifty=0cm,%
    scale=1.00,%
    rotation=0,%
    densityx=2,%
    densityy=2,%
    densityz=1,%
    dimx=4,%
    dimy=4,%
    dimz=1,%
    scalex=1,%
    scaley=1,%
    scalez=0.5,%
    anglex=0,%
    angley=90,%
    anglez=225,%
    front/.style={draw=black!50,fill=#1,opacity=0.7},%
    top/.style={draw=black!50,fill=#1,opacity=0.7},%
    right/.style={draw=black!50,fill=#1,opacity=0.7},%
    }
}

\begin{document}

\title{Sharpness-aware Low dose CT denoising using conditional generative adversarial network}


\author{Xin~Yi        \and
        Paul~Babyn 
}


\institute{X. Yi \at
              B255, Health Science Bldg, 107 Wiggins Rd, Saskatoon, SK S7N 5E5 \\
              \email{xiy525@mail.usask.ca}           
           \and
           P. Babyn \at
            Rm 1566.1, Royal University Hospital, 103 Hospital Drive Saskatoon SK S7N 0W8\\
            \email{Paul.Babyn@saskatoonhealthregion.ca}
}

\date{Received: date / Accepted: date}

\maketitle

\begin{abstract}
Low Dose Computed Tomography (LDCT) has offered tremendous benefits in radiation restricted applications, but the quantum noise  as resulted by the insufficient number of photons could potentially harm the diagnostic performance. Current image-based denoising methods  tend to produce a blur effect on the final reconstructed results especially in high noise levels. In this paper, a deep learning based approach was proposed to mitigate this problem. An adversarially trained network and a sharpness detection network were trained to guide the training process. Experiments on both simulated and real dataset shows that the  results of the proposed method have very small resolution loss and achieves better performance relative to the-state-of-art methods both quantitatively and visually.

\end{abstract}

\section{Introduction}
The use of Computed Tomography (CT) has rapidly increased over the past decade, with an estimated  80 million CT scans performed in 2015 in the United States~\cite{dose}.  Although CT offers tremendous benefits, its use has lead to significant concern regarding radiation exposure. To address this issue,  the as low as reasonably achievable (ALARA) principle has been adopted  to avoid excessive radiation dose for the patient.

Diagnostic performance should not be compromised when lowering the radiation dose. One of the most effective ways to reduce radiation dose is to reduce tube current, which has been adopted in many imaging protocols. However, low dose CT (LDCT) inevitably introduces more noise than conventional CT (convCT), which may potentially impede subsequent diagnosis or  require more advanced  algorithms for reconstruction. 
Many works have been devoted to CT denoising  with promising results  achieved by a variety of techniques, including those in the image, and sinogram domains and with iterative reconstruction techniques. One recent technique of increasing interest is deep learning (DL).

DL  has been shown to exhibit superior performance on many image related tasks,  including low level edge detection~\cite{bertasius2015deepedge}, image segmentation~\cite{yi2016lbp}, and high level vision problems including image recognition~\cite{he2016deep}, and image captioning~\cite{vinyals2015show}, with these advances now being brought into the medical domain~\cite{ chen2016low, chen2017low, kang2016deep,yang2017ct}. In this paper, we  explore the possibility of  applying generative adversarial neural net (GAN)~\cite{goodfellow2014generative} to the task of LDCT denoising.

In many image related reconstruction tasks, e.g. super resolution and inpainting, it is known that minimizing the per-pixel loss  between the output image and the ground truth alone  generate either  blurring  or make the result  visually not appealing~\cite{huang2017beyond,ledig2016photo,  zhang2017image}. We have observed the same effect in the traditional neural network based CT denoising works~\cite{ chen2016low, chen2017low, kang2016deep,yang2017ct}. The adversarial loss introduced by GAN can be treated as a driving force that can push the generated image to reside in the manifold of convCTs,  reducing the blurring effect. Furthermore, an additional sharpness detection network was also introduced to measure the sharpness of the denoised image, with focus on low contrast regions.  SAGAN (sharpness-aware generative adversarial network) will be used to denote this proposed denoising method in the remainder of the paper.

\section{Related Works}\label{related}

\textbf{LDCT Denoising}  algorithms can be broadly categorized into three groups, those conducted within the sinogram  or   image domains and  iterative reconstruction methods (which iterate back and forth across the sinogram and image domains).

The CT sinogram represents the attenuation line integrals from the radial views and is the raw projection data in the CT scan. Since the sinogram is also a 2-D signal, traditional image processing techniques have been applied for noise reduction, such as bilateral filtering~\cite{manduca2009projection}, structural adaptive filtering~\cite{balda2012ray}, etc. The filtered data can then be reconstructed to a CT image with methods like filtered back projection (FBP). Although the statistical property of the noise can be well characterized, these methods require the availability of the raw data which is not always accessible. In addition, by application of edge preservation smoothing operations (bilateral filtering), small edges would inevitably be filtered out and  lead to loss of structure and spatial resolution  in the reconstructed CT image.

Note that the above method only performs a single back projection to reconstruct the original image. Another stream of works  performs an additional forward projection, mapping the reconstructed image to the sinogram domain by modelling the acquisition process. Corrections can be made by iterating the forward and backward process. This methodology is referred as model-based iterative reconstruction (MBIR). Usually, MBIR methods  model scanner geometry and physical properties of the imaging processing, e.g. the photon counting statistics and the polychromatic nature of the source x-ray~\cite{beister2012iterative}. Some works  add prior object information  to the model to regulate the reconstructed image, such as total variation minimization~\cite{tian2011low,zhu2010duality}, Markov random fields based roughness or a sparsity penalty~\cite{bouman1993generalized}. Due to its iterative nature, MBIR models tend to consume excessive computation time for the reconstruction. There are works that are trying to accelerate the convergence behaviour of the optimization process, for example, by variable splitting of the data fidelity term~\cite{ramani2012splitting} or by combining Nesterov's momentum with ordered subsets method~\cite{kim2015combining}.

To employ the MBIR method, one also has to have access to the raw sinogram data. Image-based deniosing methods do not have this limitation. The input and output are both images. Many of the denoising methods for LDCT are borrowed from natural image processing field, such as Non-Local means~\cite{buades2005non} and BM3D~\cite{dabov2007image}. The former computes the weighted average of similar patches in the image domain while the latter is computed in a transform domain. Both methods assume the redundancy of image information. Inspired by these two seminal works, many applications have emerged  applying them into LDCTs~\cite{chen2009bayesian, chen2012thoracic,  green2016efficient, ha2015low,ma2011low,zhang2014statistical, zhang2015statistical}.  Another line of work focuses on compressive sensing, with the underlying assumption that every local path can be represented as a sparse combination of a set of bases. In the very beginning, the bases are from some generic analytic transforms, e.g. discrete gradient, contourlet~\cite{po2006directional}, curvelet~\cite{candes2002recovering}. Chen et al. built a prior image constrained compressed sensing (PICCS) algorithm for dynamic CT reconstruction under reduced views based on discrete gradient transform~\cite{lubner2011reduced}.  It has been  found that these transforms are very sensitive to both true structures and noise. Later on, bases learned directly from the source images were used with promising results. Algorithms like K-SVD~\cite{aharon2006rm} have made dictionary learning very efficient and has inspired many applications in the medical domain~\cite{chen2013improving, chen2014artifact, li2012efficient, lubner2011reduced, xu2012low}.

Convolutional neural network (CNN) based methods have recently achieved great success in image related tasks. Although its origins can be traced back to the 1980s, the resurgence of CNN can be greatly attributed to  increased computational power and  recently introduced techniques for efficient training of deep networks, such as BatchNorm~\cite{ioffe2015batch}, Rectifier linear units~\cite{glorot2011deep} and residual connection~\cite{he2016deep}.  Chen et al.~\cite{chen2016low} first used CNN to denoise CT images by learning a patch-based neural net and later on refined it with a encoder and decoder structure for end-to-end training~\cite{chen2017low}.   Kang et al.~\cite{kang2016deep} devised a 24 convolution layer net with by-pass connection and contracting path for denoising but instead of mapping in the image domain, it performed  end-to-end training in the  wavelet domain. Yang et al.~\cite{yang2017ct} adopted perceptual loss into the training, which measures the difference of the processed image and the ground truth in a high level feature space projected by a pre-trained CNN. Suzuki et al.~\cite{suzuki2017neural} proposed to use a massive-training artificial neural network (MTANN) for CT denoising. The network accepts local patches of the LDCT and regressed to the center value of the  corresponding patch of the convCT.

\textbf{Generative adversarial network} was first introduced in 2014 by Goodfellow et al.~\cite{goodfellow2014generative}.  It is a generative model trying to generate real world images by employing a min-max optimization framework where two networks (Generator  $G$ and Discriminator  $D$) are trained against each other. $G$ tries to synthesize real appearing images from random noise whereas $D$ is trying to distinguish between  the generated and real images. If the Generator  $G$ get sufficiently well trained, the Discriminator  $D$ will eventually be unable to tell if the generated image is  fake or not. 

The original setup of GAN  does not contain any constraints to control what modes of data it can generate. However, if auxiliary information were  provided during the generation, GAN can be driven to output images with specific modes. GAN in this scenario is usually referred as conditional GAN (cGAN) since the output is conditioned on additional information. Mirza et al. supplied class label encoded as one hot vector to generate MINIST digits~\cite{mirza2014conditional}. Other works have exploited the same class label information but with different network architecture~\cite{odena2016conditional, odena2016semi}. Reed et al. fed GAN with text descriptions and object locations~\cite{reed2016generative,reed2016learning}. Isola et al. proposed to do a image to image translation with GAN by directly supplying the GAN with images~\cite{isola2016image}. In this framework, training images must be aligned image pairs. Later on, Zhu et al. relaxed this restriction by introducing the cycle consistency loss so that images can be translated between two sets of unpaired samples~\cite{zhu2017unpaired}. But as also mentioned in their paper, the paired training remains the upper bound. Pathak et al. generated missing image patches conditioned on the surrounding image context~\cite{pathak2016context}. Sangkloy et al. generated images constrained by the sketched boundaries and sparse colour strokes~\cite{sangkloy2016scribbler}. Shrivastava et al. refined  synthetic images with GAN trying to narrowing the gap between the synthetic images and real image~\cite{shrivastava2016learning}. Walker et al. adopted a cGAN by conditioning on the predicted future pose information to synthesize future frames of a video~\cite{walker2017pose}.

 Two works have also applied cGAN for CT denoising. In both their works, together with ours, the denoised image is generated by conditioning on the low dose counterparts. Wolterink et al. employed a vanilla cGAN where the generator was a 7 layers all convolutional network and the discriminator is a network  to differentiate the real and denoised cardiac CT using cross entropy  loss as the objective function~\cite{wolterink2017generative}. Yang et al. adopted Wasserstein distance for the loss of the discriminator and incorporated perceptual loss to ensure visual similarity~\cite{yang2017low}. Using Wasserstein distance was claimed to be beneficial at stabilizing the training of GAN but not claimed to generate images of better quality~\cite{arjovsky2017wasserstein, gulrajani2017improved}.   Our work differs  in many ways and we would like to highlight some key points here. First, in our work,  the generator used a U-Net style network with residual components and is deeper than the other two works. The superiority of the proposed architecture in retaining small details was shown in our simulated noise experiment.  Second, our discriminator differentiates patches rather than full images which makes the resulted network have fewer parameters and applicable to arbitrary size image. Third, CT scans of a series of dose levels and anatomic regions were evaluated in this for the generality assessment. Noise and artifacts differs throughout the body. Our work showed that a singe network could potentially denoise all anatomies.   Finally, a sharpness loss was introduced to ensure the final sharpness of the image and the faithful reconstruction of low contrast regions.

\textbf{Sharpness Detection}
The sharpness detection network should be sensitive to low contrast regions. Traditional methods based on local image energy have intrinsic limitations which is the sensitivity to both the blur kernel and the image content. Recent works have proposed more sophisticated measures by exploiting the statistic differences of specific properties of blur and sharp region, e.g. gradient~\cite{shi2014discriminative}, local binary pattern~\cite{yi2016lbp}, power spectrum slope~\cite{shi2014discriminative, vu2012spectral}, Discrete Cosine Transform (DCT) coefficient~\cite{golestaneh2017spatially}. Shi et al. used sparse coding to decompose local path and quantize the local sharpness with the number of reconstructed atoms~\cite{shi2015}. There is research that tries to directly estimate the blur kernel but the estimated maps tend to be very coarse and the optimization process is very time consuming~\cite{zhu2013estimating, chakrabarti2010analyzing}.  There are other works that can produce a sharpness map, such as in depth map estimation~\cite{zhuo2011defocus}, or blur segmentation~\cite{tang2016spectral}, but the depth map is  not necessarily corresponding to the amount of sharpness and they tend to highlight blurred edges or insensitive to the change of small amount of blur.  In this work, we adopted the method of~\cite{yi2016lbp} given its sensitivity  to sharp low contrast regions. Detailed description can be found in section~\ref{sec:sharp} and~\ref{sharp}.

The rest of the paper is organized as follows. The proposed method is described in section~\ref{method}. Experiments and results are presented in section~\ref{exp} and~\ref{result}. Discussion of the potential of the proposed method is in~\ref{discussion} with conclusion drawn in section~\ref{conclusion}.

\section{Methods}\label{method}

\begin{figure}[htp]
\centering
\resizebox{\textwidth}{!}{
\begin{tikzpicture}
\node (input)[inner sep=0,label={[align=center]below:\tiny LDCT\\ $x$}] at (0,0){
    \tikz{\node (im)[anchor=south west,inner sep=0,opacity=0.5] at (0,0){\includegraphics[width=1.6cm]{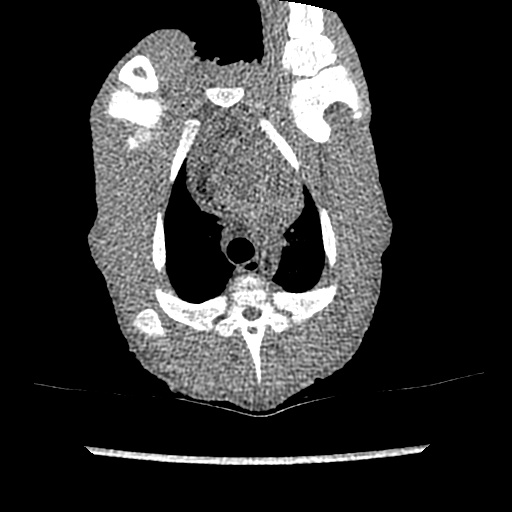}};
    \draw[step=0.2cm,draw=black,fill=red!30,very thin,opacity=0.7] (0,0) grid (1.6,1.6) rectangle (0,0);}
 };  

\node[scale=0.8,rectangle,fill=gray!20,label=left:G,node distance=10pt,rotate=90, right= 0.5cm of input, anchor=north](G) { \tikz[every node/.append style={scale=0.3}]{\pic[solid](net1) {net};}};

\node (input)[inner sep=0,label={[align=center]below:\tiny Virtual convCT\\ $\hat{y}$},right= of G.center ](output){\tikz{
    \node (im)[anchor=south west,inner sep=0,opacity=0.5] at (0,0){\includegraphics[width=1.6cm]{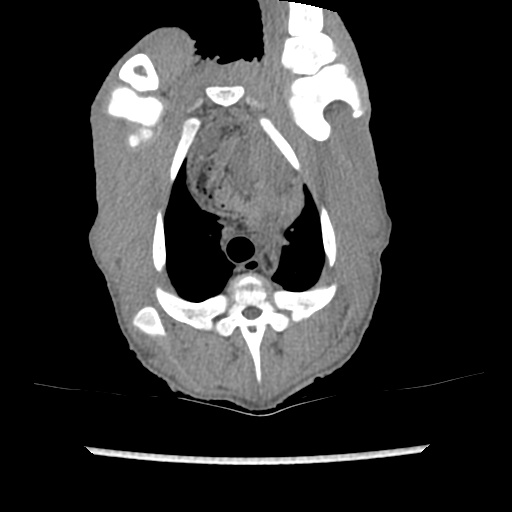}};
    \draw[step=0.2cm,draw=black,fill=blue!30,very thin,opacity=0.7] (0,0) grid (1.6,1.6)rectangle (0,0);}
};  

%


\node(inputset2)[rectangle,fill=gray!10,minimum width=50pt, right=15pt of output]{\tikz{
    	\node(cube2)[solid,scale=0.25,yshift=0.4cm,xshift=0.4cm,label={[align=center]below:\tiny Virtual convCT\\$\hat{y}$}]{\tikz{\mycube{blue!30};}};
	\node(cube3)[solid,scale=0.25,label={[align=center]below:\tiny convCT\\ $y$},right=15pt of cube2]{\tikz{\mycube{YellowGreen!40};}};
	}};
\node[scale=0.8,rectangle,fill=gray!20,label=left:  S,node distance=10pt,rotate=90, right= 0.5cm of inputset2, anchor=north](S2) { \tikz[every node/.append style={scale=0.3}]{\pic[solid](net2) {net};}};
\draw[->,>=stealth](inputset2) -- (S2)--++(30pt,0) node[text width=1.2cm,anchor=west]{\tiny Same Sharpness? \\$\mathcal{L}_{sharp}(G)$};

\node(inputset1)[rectangle,fill=gray!10,minimum width=50pt, above=15pt of inputset2]{\tikz{
    	\node(cube2)[solid,scale=0.25,yshift=0.4cm,xshift=0.4cm]{\tikz{\mycube{red!20};}};
	\node(cube1)[solid,scale=0.25,label={[align=center]below:\tiny Virtual Pair\\$(x, \hat{y})$}]{\tikz{\mycube{blue!30};}};
	\node(cube4)[solid,scale=0.25,yshift=0.4cm,xshift=0.4cm,right=15pt of cube1]{\tikz{\mycube{red!20};}};
	\node(cube3)[solid,scale=0.25,label={[align=center]below:\tiny Real Pair\\ $(x,y)$},right=15pt of cube1]{\tikz{\mycube{YellowGreen!40};}};
	}};
\node[scale=0.8,rectangle,fill=gray!20,label=left: D,node distance=10pt,rotate=90, right= 0.5cm of inputset1, anchor=north](D2) { \tikz[every node/.append style={scale=0.3}]{\pic[solid](net2) {net};}};
\draw[->,>=stealth](inputset1) -- (D2)--++(30pt,0) node[text width=1cm,anchor=west]{\tiny Virtual or Real Pair? \\$\mathcal{L}_{adv}(G,D)$};

\node(inputset3)[rectangle,fill=gray!10,minimum width=50pt, below=15pt of inputset2]{\tikz{
    	\node(cube2)[solid,scale=0.25,yshift=0.4cm,xshift=0.4cm,label={[align=center]below:\tiny Virtual convCT\\ $\hat{y}$}]{\tikz{\mycube{blue!30};}};
	\node(cube3)[solid,scale=0.25,label={[align=center]below:\tiny convCT\\ $y$},right=15pt of cube2]{\tikz{\mycube{YellowGreen!40};}};
	}};
\draw[->,>=stealth](inputset3) --++(110pt,0) node[text width=1.2cm,anchor=west]{\tiny Same Content? \\$\mathcal{L}_{L_1}(G)$};

\begin{scope}[on background layer]
    \draw[](input) -- (output);
    \draw[](output.east) -++(10pt,0) |- (inputset1.west);
    \draw[](output.east) -- (inputset2.west);
    \draw[](output.east) -++(10pt,0) |- (inputset3.west);
\end{scope}
\end{tikzpicture}
}
\caption{Overview of SAGAN. G is the generator that is responsible for the denoising. D is the discriminator employed to discriminate the virtual and real image pairs. S  is a sharpness detection network and  used to compare between the sharpness of the generated and real image. The system accepts the LDCT $x$ and convCT $y$ as the input and outputs virtual convCT (noise removed) $\hat{y}$.}
\label{overview}
\end{figure}
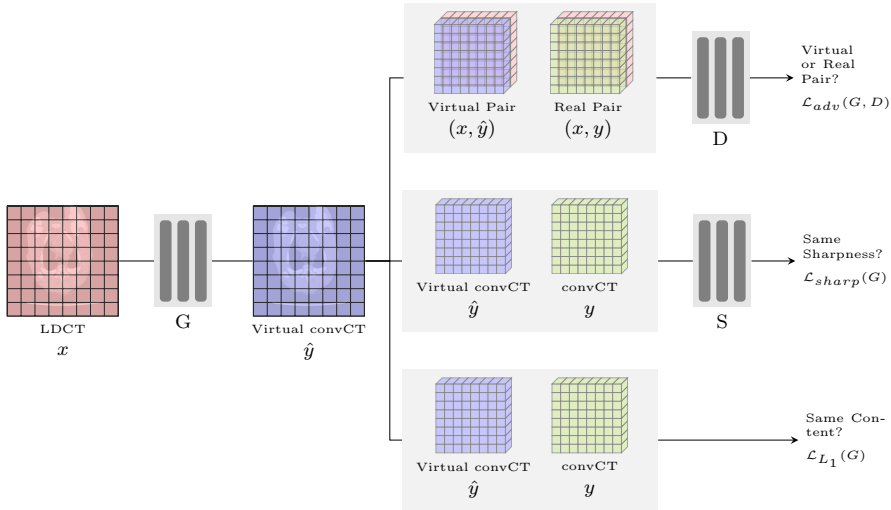

\subsection{Objective}
As shown in Figure~\ref{overview}, SAGAN consists of three networks, the generator G, discriminator D and the sharpness detection network S. G learns a mapping $G:x\rightarrow \hat{y}$, where $x$ is the LDCT the generator is conditioned upon. $\hat{y}$ is the denoised CT that is expected to be as possible close as to the convCT ($y$) and we call it virtual convCT here. D's objective is to differenciate the virtual image pair ($x, \hat{y}$) from the real one ($x, y$). Note that the input to D is not just the virtual ($\hat{y}$) and real convCT ($y$), but also LDCT ($x$). $x$ is concatenated to both $y$ and $\hat{y}$ and is served as additional information for D to rely on  so that D can penalize the mismatch. In simpler term, G tries to synthesize a virtual convCT that can fool D whereas D tries to not get fooled. The training of G against D forms the adversarial part of the objective, which can be expressed as 
\begin{equation*}
\mathcal{L}_{adv}(G,D) =\mathbb{E}_{x,y\sim p_{data}(x,y)}[(D(x,y)-1)^2] + 
					\mathbb{E}_{x\sim p_{data}(x)}[D(x,\hat{y})^2], 
\end{equation*}

G is trying to minimize the above objective whereas D is trying to maximize it. We adopt the least square loss instead of cross entropy loss in the above formulation because the least square loss tend to generate better images~\cite{mao2016least}. This loss is usually accompanied by traditional pixel-wise loss to encourage data fidelity for G, which can be expressed as 
\begin{equation*}
\mathcal{L}_{L_1}(G) =\mathbb{E}_{x,y\sim p_{data}(x,y)} [||y-\hat{y} ||_{L_1}],
\end{equation*}

Moreover, we proposed a sharpness detection network S to explicitly evaluate the denoised image's sharpness. The generator now not only has to fool the discriminator by generating a image with similar content  to the real convCT in a $L_1$ sense, but also has to generate a similar sharpness map as close as to the real convCT. With $S$ denoting the mapping from the input to the sharpness map, the sharpness loss  can be expressed as:
\begin{equation*}
\mathcal{L}_{sharp}(G) =\mathbb{E}_{x,y\sim p_{data}(x,y)}[|| S(\hat{y}) - S(y)  ||_{L_2}],
\end{equation*}

Combining these three losses together, the final objective of SAGAN is
\begin{equation*}
\mathcal{L}_{SAGAN} =\text{arg} \min_G \max_D (\mathcal{L}_{adv}(G,D)  + \lambda_1 \mathcal{L}_{L_1}(G)+ 
					\lambda_2 \mathcal{L}_{sharp}(G)),
\end{equation*}
where $\lambda_1$ and $\lambda_2$ are the weighting terms to balance the losses. Note that in the traditional GAN formulation, the generator is also conditioned on random noise $z$ to produce stochastic outputs. But in practice, people have found that the adding of noise in the conditional setup like ours tends to be not effective~\cite{isola2016image,mathieu2015deep}. Therefore, in the implementation we have discarded $z$ so that the network only produces deterministic output.

\subsection{Network Architecture}\label{pp}

\begin{figure*}[tbp]
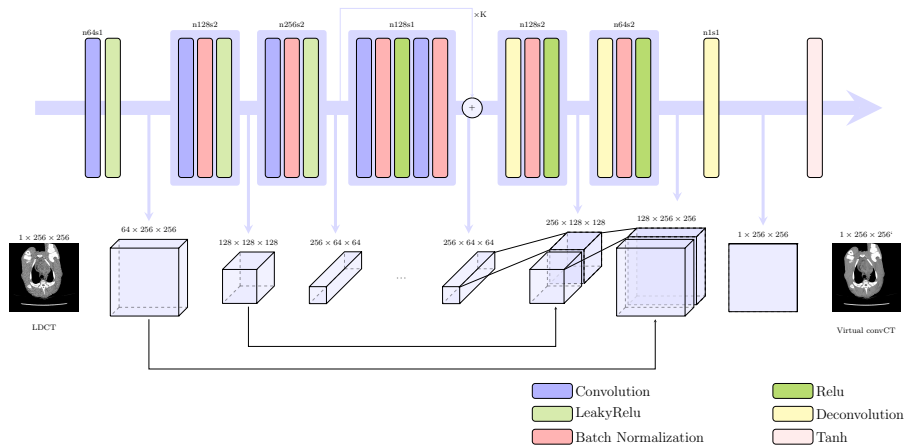

\centering
\resizebox{\textwidth}{!}{
\begin{tikzpicture}

\begin{scope}[yshift=-4cm]
\node (input)[inner sep=0,label=above:$1\times256\times256$] at (0,0){\includegraphics[width=2.4cm]{./0000520.jpg}};  
\node[right=of input,label=above:$64\times256\times256$](f1){\tikz[scale=0.4]\parapp{6}{2}{6}{blue!10}; };

\node[right=of f1,label=above:$128\times128\times128$](f2){\tikz[scale=0.4]\parapp{3}{4}{3}{blue!10}; };

\node[right=of f2,label=above:$256\times64\times64$](f3){\tikz[scale=0.4]\parapp{1.5}{8}{1.5}{blue!10}; };

\node[right= of f3,](dots){$\cdots$};

\node[right= of dots,label=above:$256\times64\times64$](f4){\tikz[scale=0.4]\parapp{1.5}{8}{1.5}{blue!10}; };

\node[right=of f4,yshift=0.7cm,xshift=0.7cm,label=above:$256\times128\times128$](f55){\tikz[scale=0.4]\parappdash{3}{4}{3}{blue!15}; };
\node[right=of f4](f5){\tikz[scale=0.4]\parapp{3}{4}{3}{blue!10}; };

\node[right=of f5,yshift=0.4cm,xshift=0.4cm,label=above:$128\times256\times256$](f66){\tikz[scale=0.4]\parappdash{6}{2}{6}{blue!15}; };
\node[right=of f5](f6){\tikz[scale=0.4]\parapp{6}{2}{6}{blue!10}; };

\node[right=of f6,label=above:$1\times256\times256$](f7){\tikz[scale=0.4]\parapp{6}{0}{6}{blue!10}; };

\node [right=of f7,label=above:$1\times256\times256$`](output) {\includegraphics[width=2.4cm]{./0000520_full.jpg}};  

\path[](input) -- coordinate[midway,yshift=6cm](p0) (f1)  ;
\path[](f1) -- coordinate[midway,yshift=6cm](p1) (f2)  ;
\path[](f2) -- coordinate[midway,yshift=6cm](p2) (f3)  ;
\path[](f3) -- coordinate[midway,yshift=6cm](p3) (f4)  ;
\path[](f4) -- coordinate[midway,yshift=6cm](p4) (f5)  ;
\path[](f5) -- coordinate[midway,yshift=6cm](p5) (f6)  ;
\path[](f6) -- coordinate[midway,yshift=6cm](p6) (f7)  ;
\path[](f7) -- coordinate[midway,yshift=6cm](p7) (output)  ;

\draw[->,>=stealth] (f1.south) -- ++(0,-50pt) -| (f6.south);
\draw[->,>=stealth] (f2.south) -- ++(0,-40pt) -| (f5.south);

\draw[] ($(f4.north east)+(-4pt,-4pt)$) -- ($(f5.north east)+(-18pt,16pt)$);
\draw[] ($(f4.north east)+(-39pt,-39pt)$) -- ($(f5.north east)+(-35pt,-1pt)$);

\draw[] ($(f55.north east)+(-4pt,-4pt)$) -- ($(f6.north east)+(-60pt,8pt)$);
\draw[] ($(f55.north east)+(-41pt,-41pt)$) -- ($(f6.north east)+(-69pt,-1pt)$);
\draw[<-,shorten <=15pt,blue!15,line width=3pt,>=stealth](f1.north) -- ++(0,125pt);
\draw[<-,shorten <=15pt,blue!15,line width=3pt,>=stealth](f2.north) -- ++(0,140pt);
\draw[<-,shorten <=15pt,blue!15,line width=3pt,>=stealth](f3.north) -- ++(0,140pt);
\draw[<-,shorten <=15pt,blue!15,line width=3pt,>=stealth](f4.north) -- ++(0,140pt);
\draw[<-,shorten <=35pt,blue!15,line width=3pt,>=stealth]($(f5.north)+(22pt,0)$) -- ++(0,140pt);
\draw[<-,shorten <=35pt,blue!15,line width=3pt,>=stealth]($(f6.north)+(22pt,0)$)-- ++(0,135pt);
\draw[<-,shorten <=15pt,blue!15,line width=3pt,>=stealth](f7.north) -- ++(0,130pt);

\node[below =10pt of input]{LDCT};
\node[below =10pt of output]{Virtual convCT};

\end{scope}

\begin{scope}[]
    \node[conv,draw,rotate=90, anchor=center,label=right:n64s1](conv1)at(p0) {};
    \node[lrelu,draw,rotate=90, anchor=center,yshift=-20pt](lrelu1) at(p0) {};
    \node[set,inner sep=8pt,rotate=90, anchor=center,label=right:n128s2](set1) at (p1){
    \tikz{
    	\pic(b1) at (0,0) {block1};
    	}
    };
    \node[set,inner sep=8pt,rotate=90, anchor=center,label=right:n256s2](set2) at (p2){
    \tikz{
    	\pic(b2) at (0,0) {block1};
    	}
    };
    
    \node[set,inner sep=8pt,rotate=90, anchor=center,label=right:n128s1](set3) at(p3){
    \tikz{
    	\pic(b3) at (0,0) {block2};
    	}
    };
    
    \node[circle,draw,inner sep=0,minimum size=20pt,fill=blue!5,rotate=90, anchor=center,yshift=40pt](plus) at(p4){+};
    \draw[->, blue!15,thick,>=stealth](set3.north) -- ++(-8pt,0) -- ++(0pt,100pt) -| (plus.east) node [midway, below,xshift=8pt,black] {$\times$K};
    \node[set,inner sep=8pt,rotate=90, anchor=center,yshift=-20pt,label=right:n128s2](set4) at(p4){
    \tikz{
    	\pic(b4) at (0,0) {block3};
    	}
    };
    \node[set,inner sep=8pt,rotate=90, anchor=center,yshift=-25pt,label=right:n64s2](set5) at(p5){
    \tikz{
    	\pic(b5) at (0,0) {block3};
    	}
    };
    
    \node[dconv,draw,rotate=90, anchor=center,label=right:n1s1](deconv1)at(p6) {};
    \node[tanh,draw,rotate=90, anchor=center](tanh)at(p7) {};
\end{scope}

\begin{scope}[on background layer]
    \draw[line width=0.5cm,shorten <=-50pt,shorten >=-60pt,->,>=stealth, blue!15](conv1.north) -- (tanh.south);
\end{scope}

\begin{scope}[yshift=-8cm,xshift=1.5*\textwidth,node distance=8pt,font=\Large]
    \node[conv,draw,label=right:Convolution,minimum height=15pt,minimum width=40pt](capconv)at(0,0) {};
    \node[lrelu,draw, below=of capconv,label=right:LeakyRelu,minimum height=15pt,minimum width=40pt](caplrelu){};
    \node[bn,draw, below=of caplrelu,label=right:Batch Normalization,minimum height=15pt,minimum width=40pt](capbn){};
    \node[relu,draw, right=200pt of capconv,label=right:Relu,minimum height=15pt,minimum width=40pt](caprelu){};
    \node[dconv,draw, right=200pt of caplrelu,label=right:Deconvolution,minimum height=15pt,minimum width=40pt](capdeconv){};
    \node[tanh,draw, right=200pt of capbn,label=right:Tanh,minimum height=15pt,minimum width=40pt](captanh){};

\end{scope}

%

\end{tikzpicture}

}
\caption{Proposed generator of the SAGAN. The residual block in the center of the network is repeated K times and K was chosen as 9 for the experiment.}
\label{fig:generator}
\end{figure*}

\subsubsection{\textbf{Generator}}
There are several different variants of generator architecture that have been adopted in the literature for image-to-image translation tasks; the Encoder-Decoder structure~\cite{hinton2006reducing}, the U-Net structure~\cite{ronneberger2015u, isola2016image}, the Residual net structure~\cite{johnson2016perceptual} and  one for removing of rain drops (denoted as Derain)~\cite{zhang2017image}. The Encoder-Decoder structure has a bottleneck layer that requires all the information  pass through it. The information consolidated by the encoder only encrypts the global structure of the input while discarding the textured details. U-Net is similar to this architecture with a slight difference in that it adds long skip connections from encoder to decoder so that fine-grained details can be recovered~\cite{ronneberger2015u}.   The residual components, first introduced by He et al.~\cite{he2016deep} is claimed to be better for the training of very deep networks. The reason is that the short skip connection of the residual component can directly guide the gradient flow from deep layer to the shallow layer~\cite{drozdzal2016importance}. Later, we start to see many works  incorporating the residual block into the network architecture~\cite{sangkloy2016scribbler, zhang2017image, zhu2017unpaired, johnson2016perceptual} when the network gets deep. The Detain architecture and its variants~\cite{yang2017low} share a common property which is that they maintained the spatial size of the feature maps during the processing. An adverse effect of this is that the number of feature maps need to remain small to avoid consuming too much memory.

Applications like style transfer  do not require preservation of local textures and details of the content image (textures come from the style image)~\cite{johnson2016perceptual}. Therefore its rare to see long skip connections used in their network structure. However, for CT noise removal, the recovery of the underlying detail is of vital importance since the subtle structure could be a lesion that can develop into cancer. Therefore, in this work we adopted the unet256 structure~\cite{isola2016image} with long skip connections. The kernel stride is 1  for the first stage feature extraction with no downsampling. We also incorporate several layers of the residual connection in the bottle neck layers for stabilizing the training of the network. Note that the feature  of the bottleneck layers' spatial  dimension is not reduced to $1\times1$ as in the Encoder-Decoder structure to reduce the model size (similar to SegNet~\cite{badrinarayanan2015segnet}) and we do not observe any significant performance drop by doing this.    The architecture can be seen in Figure~\ref{fig:generator}. An experiment to compare the different generator architecture can be found in section~\ref{gcompare}.

\subsubsection{\textbf{Discriminator}}
The objective of the discriminator is  to tell the difference between the virtual image pair $(x,\hat{y})$ and the real image pair $(x,y)$. Here, we adopt the  PatchGAN structure from pix2pix framework~\cite{isola2016image}, where instead of classifying the whole image as real or virtual, it will focus on overlapped image patches.  By using $G$ alone with $L_1$ or $L_2$ loss, the architecture would degrade to a traditional CNN-based denoising methods. 

\subsubsection{\textbf{Sharpness detection network}}\label{sec:sharp}

Bluring of the edges are a major problem faced by many image denoising methods. 
For traditional denoising methods using non-linear filtering, the edges will be inevitably blurred no matter by averaging out neighbouring pixels or self-similar patches. It is even worse in high noise settings whereas noise can also produce some edge-like structures. Neural network based methods could also suffer  the same problem if optimizing the pixel-wise differences between the generated image and the ground truth, because the result that averages  out all possible solutions end up giving the best quantitive measure. 
The adversarial loss used introduced by the discriminator of GAN is able to output a much sharper and recognizable image from the candidates. However, the adversarial loss does not guarantee the images to be sharply reconstructed, especially for low contrast regions. 

We believe that auxiliary guided information should be provided to the generator so that it can recover the underlying sharpness of the low contrast regions from the contaminated noisy image, as similar to the frontal face predication~\cite{huang2017beyond} where the position of facial landmarks are supplied to the network. Since the direct markup of low contrast sharp regions is not practical for medical images,  an independent sharpness detection network S was trained in this work. During the training of SAGAN, the virtual convCT generated from G is sent to  S and the output sharpness map is compared with the map of the ground truth image. We compute the mean square error between the two sharpness maps and this error was back-propagated through the generator to update its weights.

\section{Experiment Setup}\label{exp}
The proposed SAGAN was applied to both simulated low dose and real low dose CT images to evaluate its effectiveness. In both settings, peak signal to noise ratio (PSNR)  and structured similarity index (SSIM)~\cite{wang2004image} were adopted as the quantitive metrics for the evaluation (using abdomen window image). The former metric is commonly used to measure the pair-wise difference of two signals whereas the SSIM is claimed to better conform  to the human visual perception. For the real dataset, the mean standard deviation of  42 smooth rectangular homogeneous regions (size of $21 \times 21$, 172.27 $\text{mm}^2$) were computed as direct measures of the noise level.

To further evaluate the general applicability of the proposed method, we selected two patient's LDCTs from the Kaggle Data Science Bowl 2017~\cite{kaggle} and applied our trained model to it. Visual results and noise levels are provided for evaluation in this case. 20 Rectangular   homogeneous regions of size $21 \times 21$ were selected for the calculation.

\subsection{Simulated Noise Dataset}\label{simdatasets}
In this dataset, 239 normal dose CT images were downloaded from the National Cancer Imaging Archive (NBIA). Each image has a size of $512\times512$ covering different parts of the human body. A fan-beam geometry was used to transform the image to the sinogram, utilizing  937 detectors and 1200 views.

 The photons hitting the detector are  usually treated as  Possion distributed. Together with the electrical noise which starts to gain prominence in low dose cases and is normally Gaussian distributed, the detected signal can be expressed as:
\begin{equation}
N \sim \text{Poisson}(N_{0}\text{exp}(-y)) + \text{Gaussian}(0,\sigma_e^2),
\label{noise}
\end{equation}
where $N_0$ is the X-ray source intensity or sometimes called blank flux, $y$ is the sinogram data, and $\sigma_e$ is the standard deviation of the electrical noise~\cite{zhang2014statistical, la2005penalized}.


The blank scan flux $N_0$ was set to be $1\times10^5, 5\times10^4, 3\times 10^4, 1\times10^4$ to simulate effect of different dose levels and the electrical noise was discarded for simplicity.  
 Since the network used is fully convolutional, the input can be of different size. Each image was further divided into four $256\times256$ sub-images to boost the size of the dataset. 700 out of the resultant 956 sub-images were randomly selected as the training set and the remaining 64 full  images were used as the test set. Some sample images are shown in the first column of Figure~\ref{fig:edgy}. Note that the simulated dose here is generally lower than that of ~\cite{chen2017low}.

\begin{table}[!t]
\caption{Detailed doses for the piglet and phantom datasets.}
\label{realoverview}
\centering
\begin{tabular}{cccccc} \toprule
Dose level					&	Full		&	50\%			&	25\%		&	10\%		&	5\%		\\ \midrule
Tube current (mAs)				&	300 		&	150			&	 75		&	30		&	15		\\  \midrule

$\text{CTDI}_{\text{vol}}$ (mGy)		&	30.83 	&	15.41		&	7.71		&	3.08		&	1.54		\\ \midrule

DLP (mGy-cm)					&	943.24 	&	471.62		&	 235.81	&	94.32	&	47.16	\\  \midrule

Effective dose (mSv)				&	14.14 	&	7.07			&	 3.54		&	1.41		&	0.71	\\  \bottomrule \vspace{8pt}
\end{tabular}
\caption*{(a) Doses used for the piglet dataset. In all 5 series, tube potential was 100 kV with 0.625 mm slice thickness. Tube currents were decreased to 50\%, 25\%, 10\%, 5\% of full dose tube current (300mAs) to obtained images with different doses. }
\bigskip
\begin{tabular}{ccc} \toprule
Scan series					&	Full		&	3.33\%		\\ \midrule
Tube current (mAs)				&	300 		&	10		\\  \midrule

$\text{CTDI}_{\text{vol}}$ (mGy)		&	26.47 	&	0.88		\\ \bottomrule \vspace{8pt}
\end{tabular}
\caption*{(b) Doses used for the Catphan 600 dataset. In both series, tube potential is 120 kV with 0.625 mm slice thickness. }

\end{table}%

%
%
%

\subsection{Real Datasets}
CT scans of a deceased piglet were obtained with a range of different doses utilizing a GE scanner (Discovery CT750 HD) using 100 kVp source potential and 0.625 mm slice thickness. A series of tube currents were used in the scanning to produce images with different dose levels, ranging from 300 mAs down to 15 mAs. The 300 mAs  served as the conventional full dose whereas the others served as low doses with tube current reductions of 50\%, 25\%, 10\% and 5\% respectively.  At each dose level we obtained 850 images of a size $512\times512$ in total. 708 of them were selected for training and 142 for testing. The training size of the real dataset was also boosted by dividing each image into four $256\times256$ sub-images, giving us 2832 images in total for training.

A CT phantom (Catphan 600) was scanned to  evaluate the spatial resolution of the reconstructed image, using 120 kVp and 0.625mm slice thickness. For this dataset, only two dose levels were used. The one with 300 mAs served as the convCT and the one with 10mAs served as the LDCT. The detailed doses is provided in Table~\ref{realoverview}.

Data Science Bowl 2017 is a challenge to detect lung cancer from LDCTs. It contains over a thousand high-resolution low-dose CT images of high risk patients. The corresponding convCTs and specific dosage level for each scan are not available.  We selected two patients' scan to evaluate the generality of the proposed SAGAN method on unseen doses.

Four experiments were conducted. In the first experiment, we evaluated the effect of the generator and the sharpness loss by using the simulated noise dataset. In the second experiment, we evaluated the spatial resolution  with the Catphan 600 dataset. In the third experiment the proposed SAGAN was applied on the piglet dataset to test its effectiveness on a wide range of real quantum noise. Finally in the last experiment, the SAGAN model trained on the piglet dataset was applied to the clinical patient data in the Data Science Bowl 2017. Two state-of-the-art methods: BM3D, K-SVD from two major line of traditional image denoising methods were selected for the comparison. For the real dataset, the available CT manufacture iterative reconstruction methods, ASIR (40\%) and VEO were also compared. All experiments on the real datasets are trained on the full range DICOM image.

\begin{figure}[!t]
\centering
\resizebox{\textwidth}{!}{
\begin{tikzpicture} [
    auto,
    line/.style     = { draw, thick, ->, shorten >=2pt,shorten <=2pt },
    every node/.append style={font=\large}
  ]
 \matrix [column sep=2mm, row sep=2mm,ampersand replacement=\&] {
		 \node (p11)[inner sep=0] at (0,0){\includegraphics[width=0.3\textwidth]{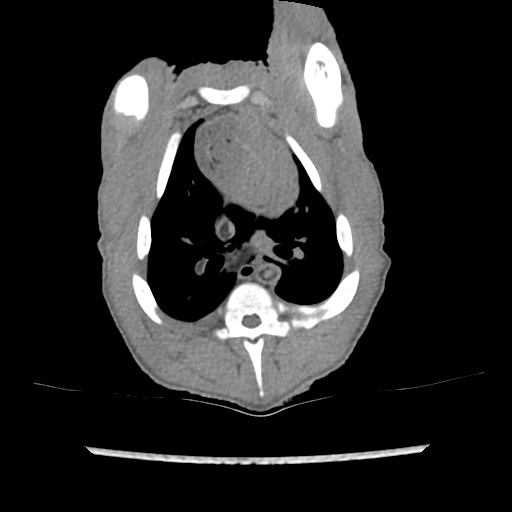}};     \&
		 \node (p12)[inner sep=0] at (0,0){\includegraphics[width=0.3\textwidth]{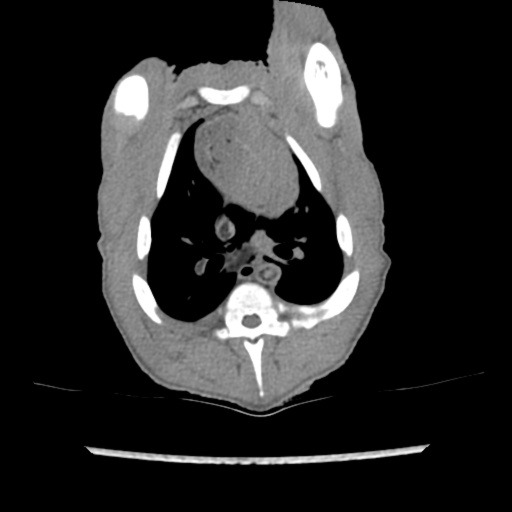}};     \&
		 \node (p13)[inner sep=0] at (0,0){\includegraphics[width=0.3\textwidth]{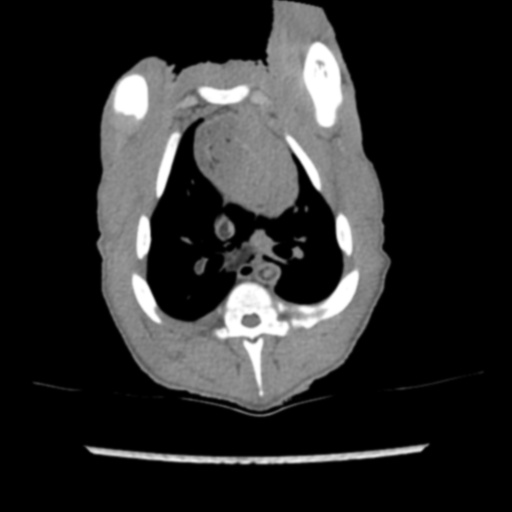}};     \&
		 \node (p14)[inner sep=0] at (0,0){\includegraphics[width=0.3\textwidth]{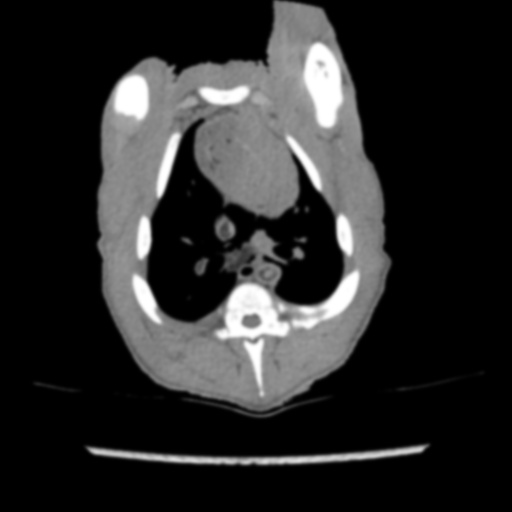}};     \&
		 \node (p15)[inner sep=0] at (0,0){\includegraphics[width=0.3\textwidth]{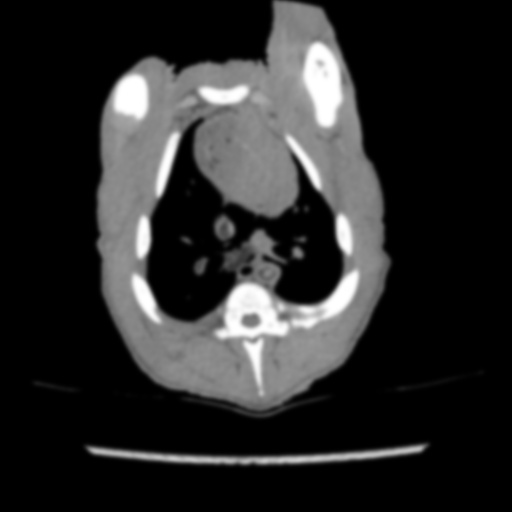}};     \&\\
 		 \node (p21)[inner sep=0] at (0,0){\includegraphics[width=0.3\textwidth]{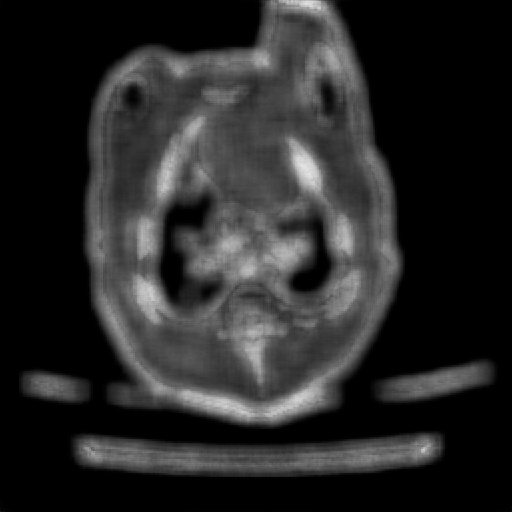}};     \&
		 \node (p22)[inner sep=0] at (0,0){\includegraphics[width=0.3\textwidth]{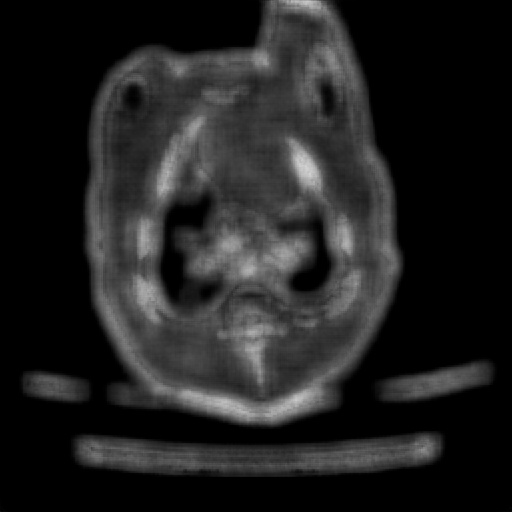}};     \&
		 \node (p23)[inner sep=0] at (0,0){\includegraphics[width=0.3\textwidth]{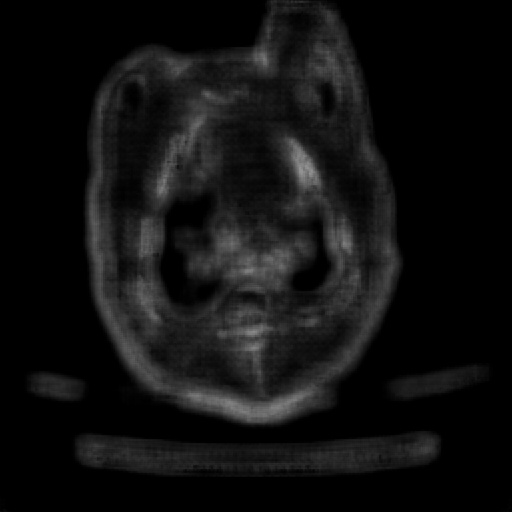}};     \&
		 \node (p24)[inner sep=0] at (0,0){\includegraphics[width=0.3\textwidth]{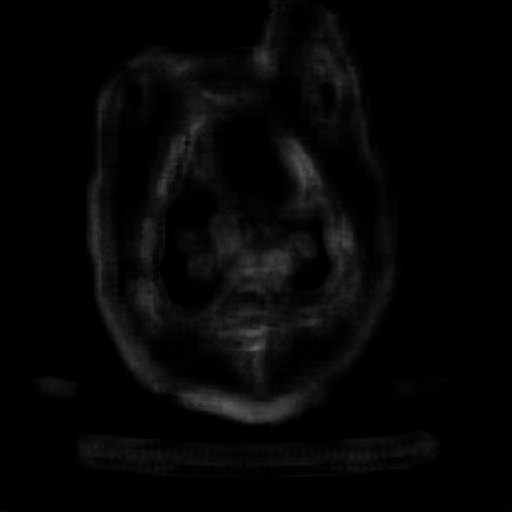}};     \&
		 \node (p25)[inner sep=0] at (0,0){\includegraphics[width=0.3\textwidth]{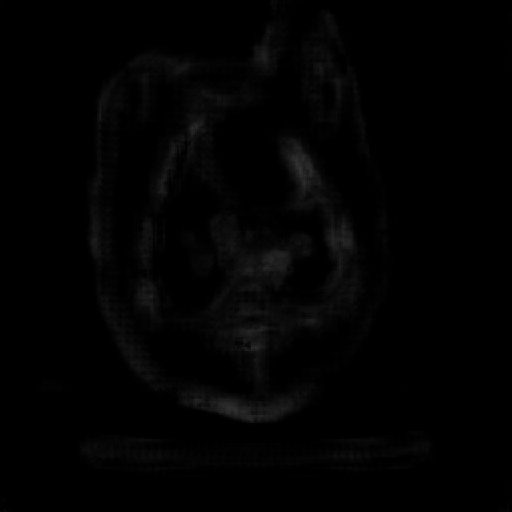}};     \&\\
                       };
  \begin{scope} [every path/.style=line]
     \node[anchor=south] at (p11.north) {convCT};
     \node[anchor=south] at (p12.north) {$\sigma=0.5$};
     \node[anchor=south] at (p13.north) {$\sigma=1.0$};
     \node[anchor=south] at (p14.north) {$\sigma=1.5$};
     \node[anchor=south] at (p15.north) {$\sigma=2.0$};

  \end{scope}
\end{tikzpicture}
}
\caption{The output of the sharpness detection network. The upper row is the convCT of a lung region selected from the piglet dataset  and its blurred versions with increasing amount of Gaussian blur ($\sigma$ shown on top). The lower row shows their corresponding sharpness map.}
\label{fig:sharpnessmap}
\end{figure}
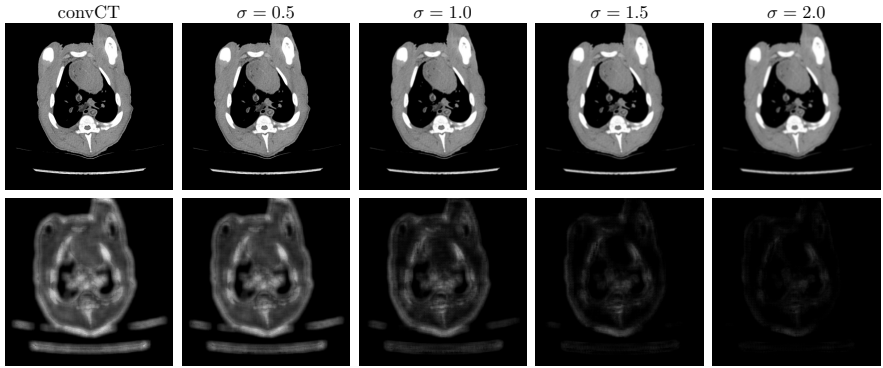

\subsection{Implementation Details}

\subsubsection{Training of SAGAN}
All the networks are trained on the Guillimin cluster of Calcul Quebec and the Cedar cluster of Compute Canada.  Adam optimizer~\cite{kingma2014adam} with $\beta_1=0.5$ was used for the optimization with learning rate  0.0001. The generator and discriminator was trained alternatively across the training with $k=1$ as similar in~\cite{goodfellow2014generative}. The implementation was based on the Torch framework. The training images have size of $256\times256$ whereas the testing is with full size $512\times512$ CT images. All the networks here are trained to 200 epochs. $\lambda_1$ was set to be 100 and $\lambda_2$ to 0.001. For the simulated dataset, one SAGAN was trained for each simulated dose. For the real dataset, one SAGAN was trained for the piglet and phantom separately, with  the  training set of different doses of piglet combined.

\subsubsection{Training of the sharpness detection network}\label{sharp}
The sharpness detection network follows the work of Yi et al.~\cite{yi2016lbp}. In that work, Yi et al. used a non-differentiable analytic sharpness metric to quantify the local sharpness of a single image. Here in this work, we trained a neural network to imitate its behaviour. To be more specific, the defocus segmentation dataset~\cite{dataset} that contains 704 defocused images was adopted for the training.   5 subimages of size $256\times256$ were sampled from the 4 corners and centre of each defocus image to boost the size of the training set. For the training of the sharpness detection network, the unet256 structure was adopted and the sharpness map created by  the local sharpness metric of~\cite{yi2016lbp} was used for regression. Adam optimizer~\cite{kingma2014adam} with $\beta_1=0.5$ was also used for the optimization with learning rate also 0.0001.   Some sample images and their sharpness map can be seen in Figure~\ref{fig:sharpnessmap}.

\section{Results}\label{result}

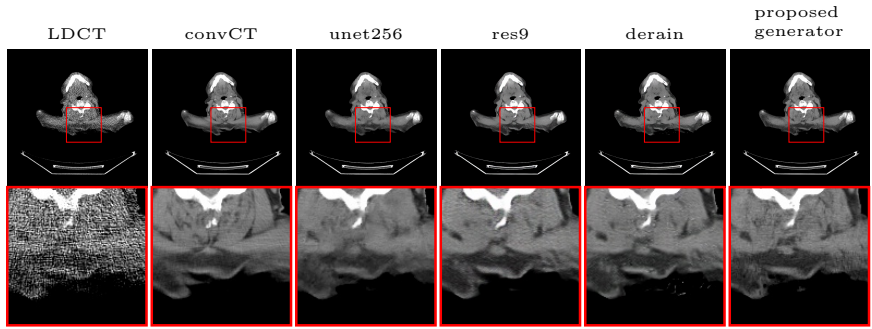
\begin{figure}[!t]
\centering
\resizebox{\textwidth}{!}{
\begin{tikzpicture} [
    auto,
    line/.style     = { draw, thick, ->, shorten >=2pt,shorten <=2pt },    every node/.append style={font=\tiny}
  ]
\newcommand{\ax}{0.55}
\newcommand{\ay}{0.55}
\newcommand{\imgnum}{0000184}
 \matrix [column sep=0.3mm, row sep=7mm,ampersand replacement=\&] {
 		 \node (p11)[inner sep=0] at (0,0){\magimage{./simulate/poisson_10000/input/\imgnum.jpg}{0.13\textwidth}{\ax}{\ay}{0.13\textwidth}};     \&
		 \node (p12)[inner sep=0] at (0,0){\magimage{./simulate/poisson_10000/target/\imgnum.jpg}{0.13\textwidth}{\ax}{\ay}{0.13\textwidth}};     \&
 		 \node (p13)[inner sep=0] at (0,0){\magimage{./simulate/poisson_10000/output_unet256/\imgnum.jpg}{0.13\textwidth}{\ax}{\ay}{0.13\textwidth}};     \& 
 		 \node (p21)[inner sep=0] at (0,0){\magimage{./simulate/poisson_10000/output_resnet9/\imgnum.jpg}{0.13\textwidth}{\ax}{\ay}{0.13\textwidth}};     \&
		 \node (p22)[inner sep=0] at (0,0){\magimage{./simulate/poisson_10000/output_zhang/\imgnum.jpg}{0.13\textwidth}{\ax}{\ay}{0.13\textwidth}};     \&
 		 \node (p23)[inner sep=0] at (0,0){\magimage{./simulate/poisson_10000/output_ssne/\imgnum.jpg}{0.13\textwidth}{\ax}{\ay}{0.13\textwidth}};     \& \\
                       };
  \begin{scope} [every path/.style=line]
     \node[anchor=south] at (p11.north) {LDCT};
     \node[anchor=south] at (p12.north) {convCT};
     \node[anchor=south] at (p13.north) {unet256};
     \node[anchor=south] at (p21.north) {res9};
     \node[anchor=south] at (p22.north) {derain};
     \node[anchor=south,text width=1cm] at (p23.north) {proposed generator};
  \end{scope}
\end{tikzpicture}
}
\caption{Evaluation of the generator architecture. LDCT is with $N_0=10000$.}
\label{fig:ga}
\end{figure}

\begin{table*}[tbp]
\caption{Comparison of different generator architecture on the simulated dataset. The input noise level in terms of PSNR and SSIM is shown in the top row.}
\label{generatorcmp}
\begin{center}
\resizebox{\textwidth}{!}{
\begin{tabular}{ccccccccc}\toprule
\multirow{2}{*}{\makecell{Generator \\Archetecture}}		&	\multicolumn{2}{c}{$N_0=10000$} 	&	\multicolumn{2}{c}{$N_0=30000$}		&	\multicolumn{2}{c}{$N_0=50000$} 	&	\multicolumn{2}{c}{$N_0=100000$} \\ \cmidrule{2-9}
									&	\makecell{PSNR\\18.3346} 	&	\makecell{SSIM\\0.7557} 			&	\makecell{PSNR\\21.6793} 	&	\makecell{SSIM	\\0.7926}&	\makecell{PSNR\\23.1568} 	&	\makecell{SSIM	\\0.8119}			&	\makecell{PSNR\\24.8987} 	&	\makecell{SSIM	\\0.8387} 	\\\midrule
unet256								&	26.2549	&	0.8384			&	27.5819	&	0.8598	&	27.9269	&	0.8646			&	28.1243	&	0.8711	\\
res9									&	25.9032	&	0.8412			&	26.7443	&	0.8549	&	27.8765	&	0.8710			&	28.8179	&	0.8877	\\
derain								&	25.8094	&	0.8376			&	26.4167	&	0.8505	&	27.1724	&	0.8562			&	27.1307	&	0.8570 	\\
proposed								&	26.6584	&	0.8438			&	27.3066	&	0.8533	&	27.8443	&	 0.8622			&	28.1718	&	0.8701	\\
\bottomrule

\end{tabular}
}
\end{center}

\end{table*}%

\subsection{Analysis of the generator architecture}\label{gcompare}
 A variety of generator architectures were evaluated, including unet256~\cite{ronneberger2015u},   res9~\cite{johnson2016perceptual}, Derain~\cite{zhang2017image}. In the analysis, only the architecture of the generator was modified. The discriminator was fixed to be the patchGAN with patch size of $70\times70$~\cite{isola2016image}. The sharpness network was not incorporated in this experiment for simplicity.

The quantitative results are shown in Table~\ref{generatorcmp}. As can be seen that the performance generally improves when lowering the noise level (increase of $N_0$) no matter what architecture has been used. For every single noise level,  the listed generators achieved comparative results since all of them  optimized the PSNR as part of their loss function. However, the visual results shown in~Figure~\ref{fig:ga} have shed some light on the architectural  differences. Comparing the unet256 with the proposed, we can see that the proposed  recovered the low contrast zoom region much sharper. It shows the benefits of maintaining the spatial size during the first stage of feature extraction.  The major difference of res9 and the proposed one is the long skip connection. We can see by comparing  results of the two that this connection can help recovering small details. As for the derain architecture, the training was not stable and  some grainy artifacts can be observed. We attribute this to the small size of the feature dimension of  the bottleneck layer (only 1 in this case) which is not  sufficient to encode the global features.

\begin{table*}[!t]
\caption{Quantitive evaluation of the sharpness-aware loss.}
\label{sharpnesscmp}
\begin{center}
\resizebox{\textwidth}{!}{
\begin{tabular}{ccccccccc}\toprule
\multirow{2}{*}{Methods}		&	\multicolumn{2}{c}{$N_0=10000$} 	&	\multicolumn{2}{c}{$N_0=30000$}		&	\multicolumn{2}{c}{$N_0=50000$} 	&	\multicolumn{2}{c}{$N_0=100000$} \\ \cmidrule{2-9}
									&	\makecell{PSNR\\18.3346} 	&	\makecell{SSIM\\0.7557} 			&	\makecell{PSNR\\21.6793} 	&	\makecell{SSIM	\\0.7926}&	\makecell{PSNR\\23.1568} 	&	\makecell{SSIM	\\0.8119}			&	\makecell{PSNR\\24.8987} 	&	\makecell{SSIM	\\0.8387} 	\\\midrule
w/o sharpness loss						&	26.6584	&	0.8438			&	27.3066	&	0.8533	&	\textbf{27.8443}	&	 \textbf{0.8622}			&	28.1718	&	0.8701	\\
w sharpness loss (SAGAN) 				&	\textbf{26.7766}	&	\textbf{0.8454}			&	\textbf{27.5257}	&	\textbf{0.8571}	&	27.7828	&	 0.8620			&	\textbf{28.2503}	&	\textbf{0.8708}	\\\midrule
BM3D								&	24.0038	&	0.8202			&	25.6046	&	0.8485	&	26.0913	&	0.8589			&	26.7598	&0.8726	\\	
K-SVD								&	21.9578	&	0.7665			&	24.0790	&	0.8167	&	25.0425	&	0.8379			&	26.0902	&	0.8620		\\	
\bottomrule

\end{tabular}
}
\end{center}

\end{table*}%

\subsection{Analysis of the sharpness loss}
\begin{figure*}
\centering
\resizebox{\textwidth}{!}{
\begin{tikzpicture} [
    auto,
    line/.style     = { draw, thick, ->, shorten >=2pt,shorten <=2pt },
    every node/.append style={font=\tiny}
  ]
   \newcommand{\ax}{0.7}
\newcommand{\ay}{0.6}
 \newcommand{\bx}{0.65}
\newcommand{\by}{0.55}
 \matrix [column sep=0.3mm, row sep=2mm,ampersand replacement=\&] {
 		 \node (p11)[inner sep=0] at (0,0){\magimage{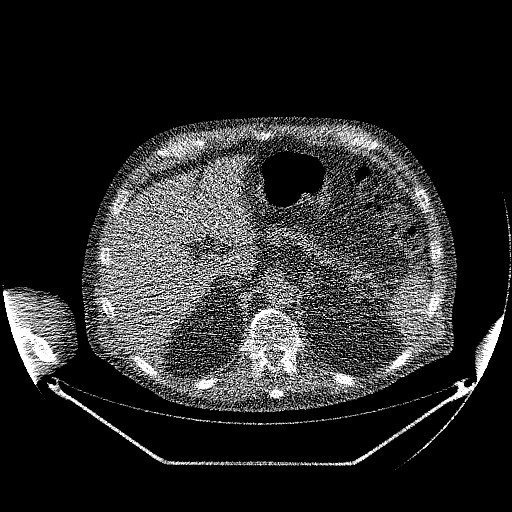}{0.13\textwidth}{0.4}{0.5}{0.13\textwidth}};     \&
		 \node (p12)[inner sep=0] at (0,0){\magimage{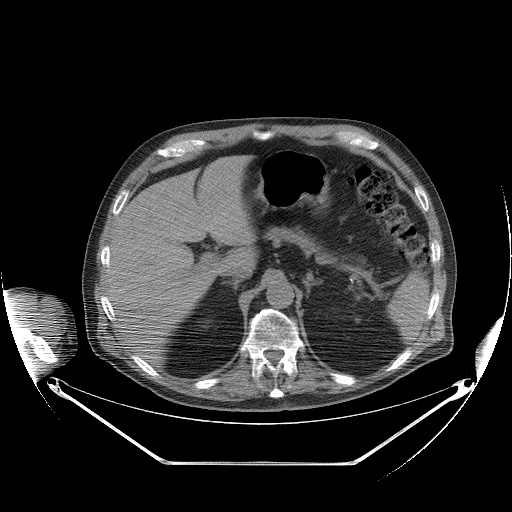}{0.13\textwidth}{0.4}{0.5}{0.13\textwidth}};     \&
 		 \node (p13)[inner sep=0] at (0,0){\magimage{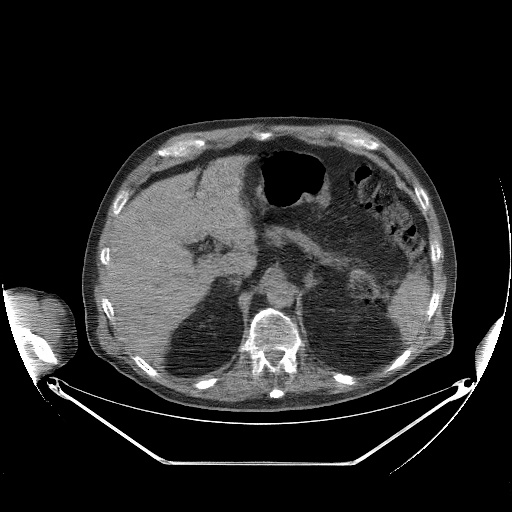}{0.13\textwidth}{0.4}{0.5}{0.13\textwidth}};     \& 
		 \node (p14)[inner sep=0] at (0,0){\magimage{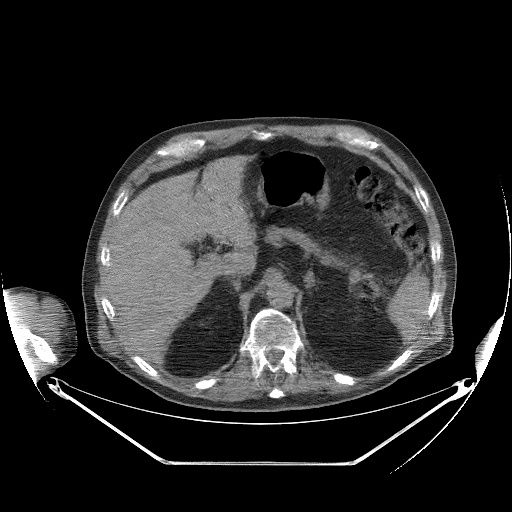}{0.13\textwidth}{0.4}{0.5}{0.13\textwidth}};     \&
		 \node (p15)[inner sep=0] at (0,0){\magimage{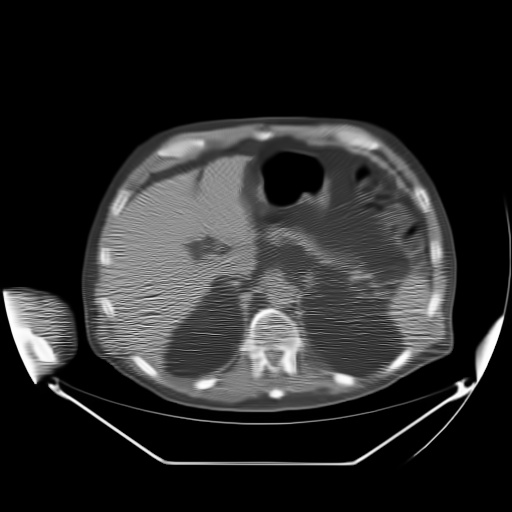}{0.13\textwidth}{0.4}{0.5}{0.13\textwidth}};     \&
		 \node (p16)[inner sep=0] at (0,0){\magimage{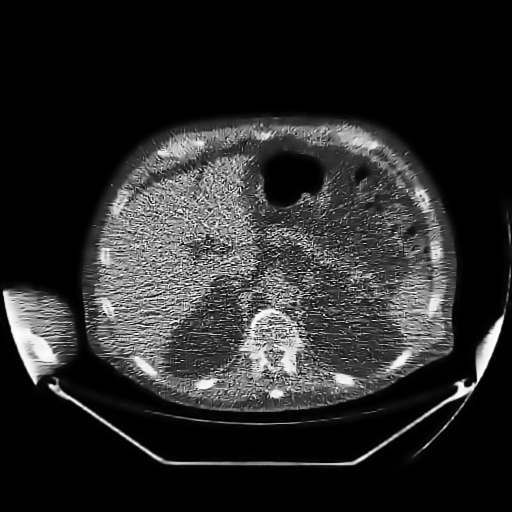}{0.13\textwidth}{0.4}{0.5}{0.13\textwidth}};     \&\\
 		 \node (p21)[inner sep=0] at (0,0){\magimage{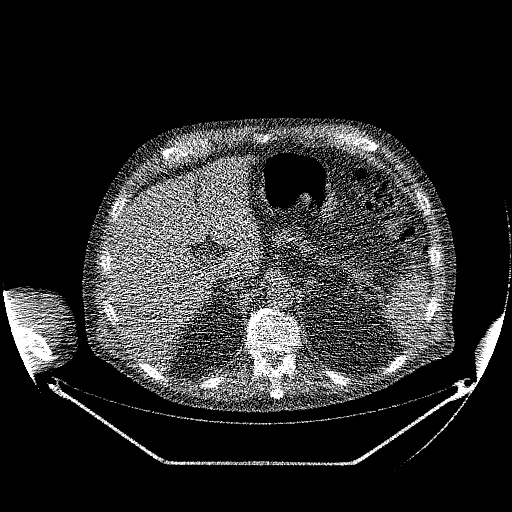}{0.13\textwidth}{\bx}{\by}{0.13\textwidth}};     \&
		 \node (p22)[inner sep=0] at (0,0){\magimage{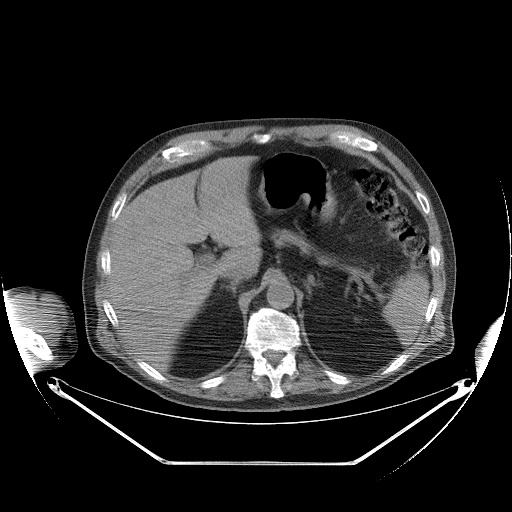}{0.13\textwidth}{\bx}{\by}{0.13\textwidth}};     \&
 		 \node (p23)[inner sep=0] at (0,0){\magimage{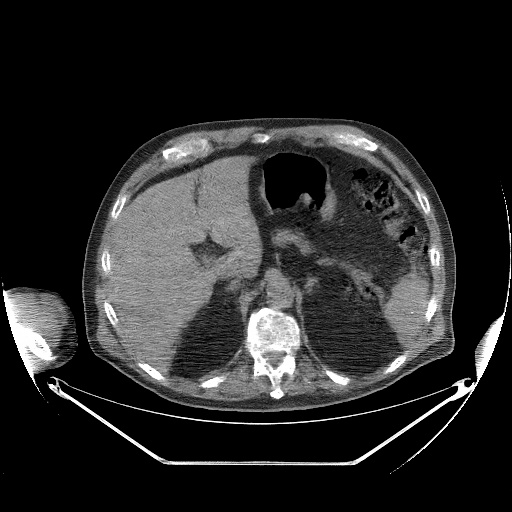}{0.13\textwidth}{\bx}{\by}{0.13\textwidth}};     \& 
		 \node (p24)[inner sep=0] at (0,0){\magimage{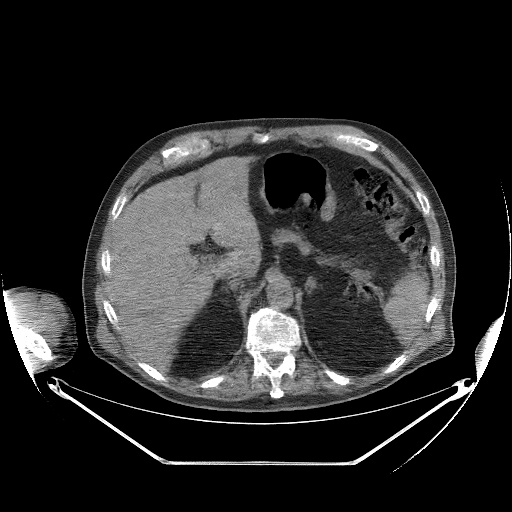}{0.13\textwidth}{\bx}{\by}{0.13\textwidth}};     \&
		 \node (p25)[inner sep=0] at (0,0){\magimage{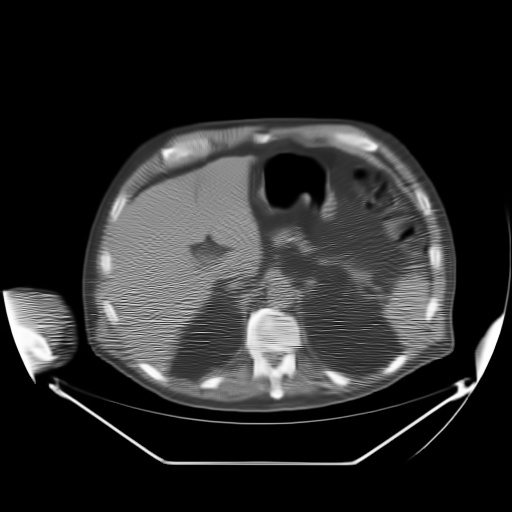}{0.13\textwidth}{\bx}{\by}{0.13\textwidth}};     \&
		 \node (p26)[inner sep=0] at (0,0){\magimage{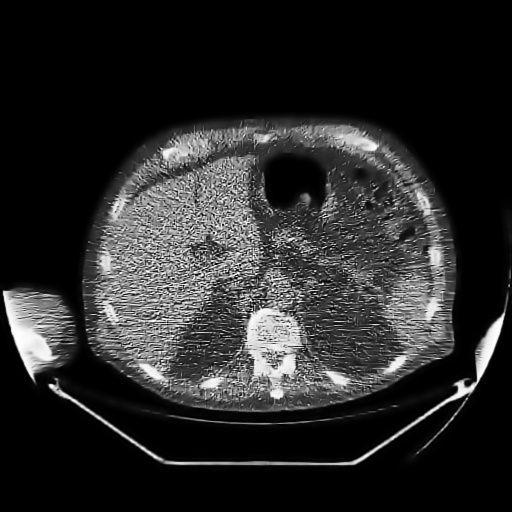}{0.13\textwidth}{\bx}{\by}{0.13\textwidth}};     \&\\
                       };
  \begin{scope} [every path/.style=line]
     \node[anchor=south] at (p11.north) {LDCT};
      \node[anchor=south] at (p12.north) {convCT};
       \node[anchor=south] at (p13.north) {w/o sharp loss};
        \node[anchor=south,text width=0.13\textwidth,align=center] at (p14.north) {w sharp loss \\(SAGAN)};
         \node[anchor=south] at (p15.north) {BM3D};
          \node[anchor=south] at (p16.north) {K-SVD};

  \end{scope}
\end{tikzpicture}
}
\caption{Visual examples to evaluate the effectiveness of the proposed sharpness loss with $N_0=10000$. Row 1 and 3 are two examples with zoomed region shown below. }
\label{fig:edgy}
\end{figure*}
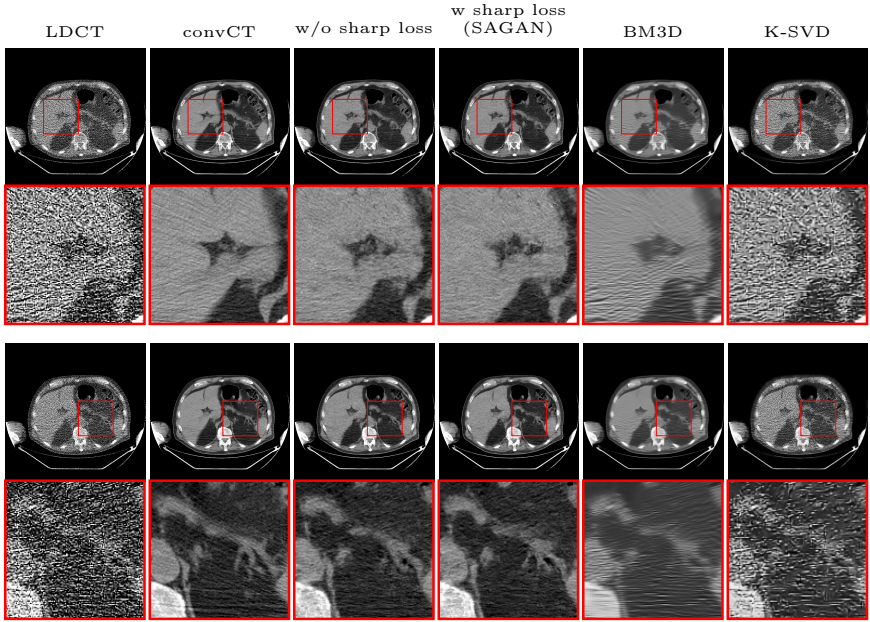

In this experiment, we evaluated the effectiveness of the sharpness loss on the denoised result. Table~\ref{sharpnesscmp} shows quantitive results  before and after applying the sharpness detection network. The values in term of PSNR and SSIM are comparable to each other with slight differences that can be explained by the competition of the data fidelity loss and the sharpness loss. However, visual examples shown in Figure~\ref{fig:edgy} clearly demonstrates that the sharpness loss excels at suppressing noises on small structures without introducing too much blurring.

\subsection{Denoising results on simulated dataset}
As can be also seen from Table~\ref{sharpnesscmp} and Figure~\ref{fig:edgy}, the performance of SAGAN in terms of PSNR and SSIM is better than BM3D and K-SVD  in all noise levels.  For the visual appearance, SAGAN is also sharper than BM3D and K-SVD and can recover more details. K-SVD could not remove all the noise and sometimes make the resultant image very blocky. For example, in the zoomed region of row 1 of Figure~\ref{fig:edgy}, the fat region of SAGAN is the sharpest among the comparators. In row 2, we have shown that the low contrast vessel can be faithfully reconstructed by SAGAN whereas missed by the other methods.  The streak artifact is another problem faced by BM3D in high quantum noise level as has already been pointed out by many works~\cite{chen2017low, kang2016deep}.  We recommend the reader to zoom in for better appreciation of the results.

\subsection{Denoising results on Catphan 600}
\begin{figure}[!b]
\centering
\resizebox{\textwidth}{!}{
\begin{tikzpicture} [
    auto,
    line/.style     = { draw, thick, ->, shorten >=2pt,shorten <=2pt },
    every node/.append style={font=\tiny}
  ]
 \newcommand{\ax}{0.4}
\newcommand{\ay}{0.3}
 \newcommand{\bx}{0.55}
\newcommand{\by}{0.3}
 \matrix [column sep=0.3mm, row sep=7mm,ampersand replacement=\&] {
 		 \node (p11)[inner sep=0] at (0,0){\magimage{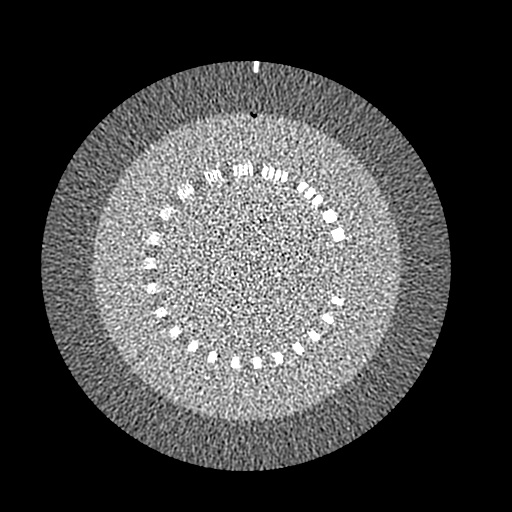}{0.13\textwidth}{\ax}{\ay}{0.13\textwidth}};     \&
		 \node (p12)[inner sep=0] at (0,0){\magimageref{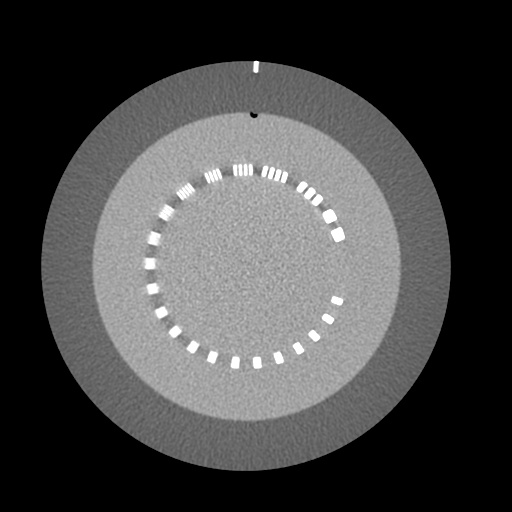}{0.13\textwidth}{\ax}{\ay}{0.13\textwidth}};     \&
 		 \node (p13)[inner sep=0] at (0,0){\magimage{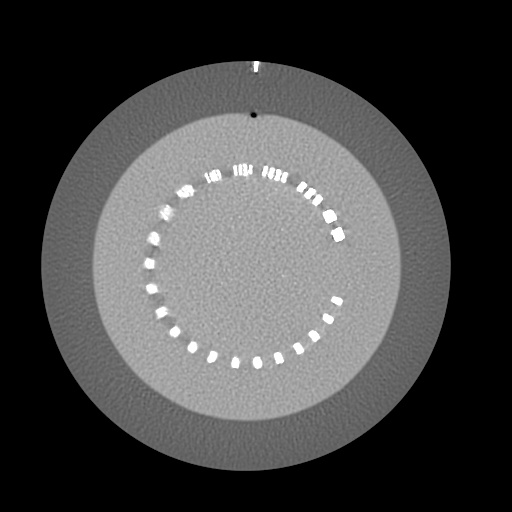}{0.13\textwidth}{\ax}{\ay}{0.13\textwidth}};     \& 
		 \node (p14)[inner sep=0] at (0,0){\magimage{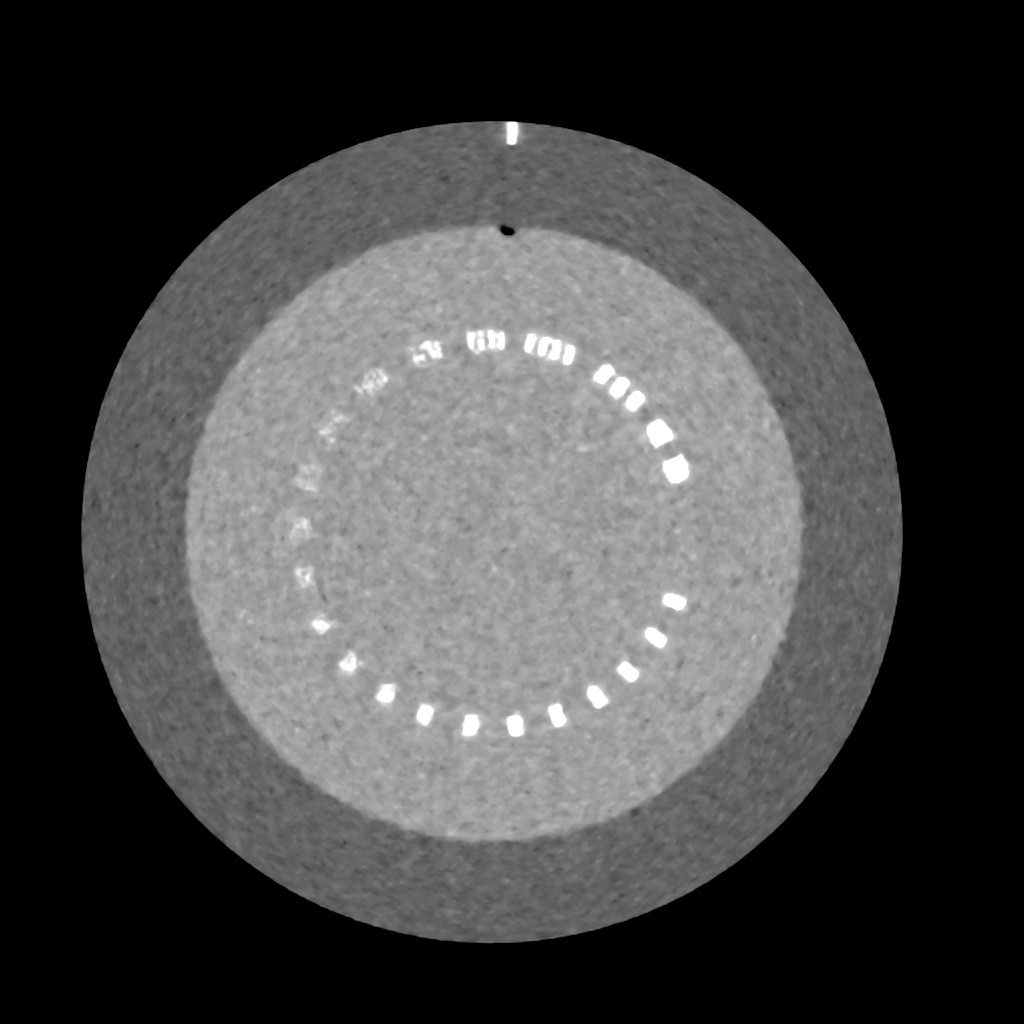}{0.13\textwidth}{\ax}{\ay}{0.13\textwidth}};     \&
		 \node (p15)[inner sep=0] at (0,0){\magimage{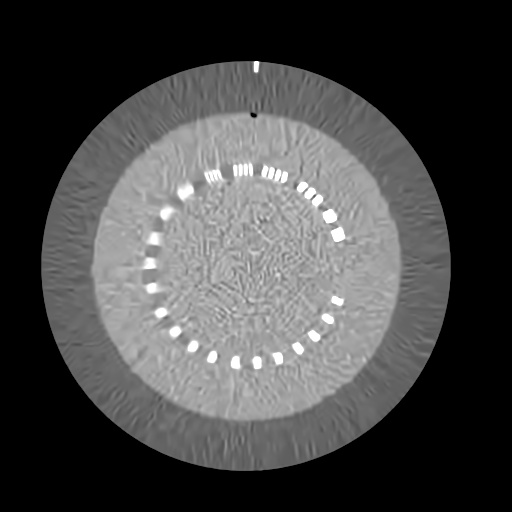}{0.13\textwidth}{\ax}{\ay}{0.13\textwidth}};     \&
		 \node (p16)[inner sep=0] at (0,0){\magimage{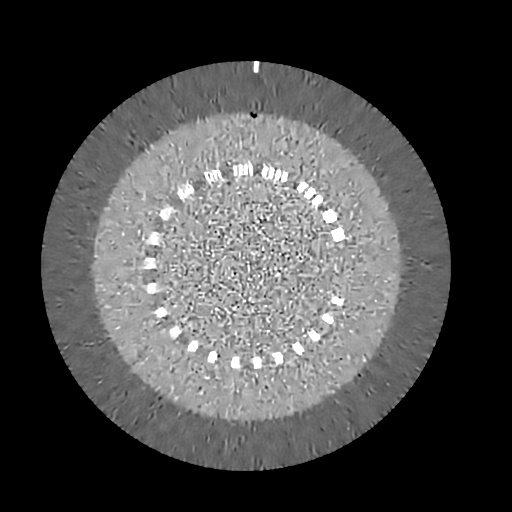}{0.13\textwidth}{\ax}{\ay}{0.13\textwidth}};     \&\\
                       };
  \begin{scope} [every path/.style=line]
     \node[anchor=south] at (p11.north) {LDCT};
      \node[anchor=south] at (p12.north) {convCT};
       \node[anchor=south] at (p13.north) {SAGAN};
        \node[anchor=south] at (p14.north) {VEO};
        \node[anchor=south] at (p15.north) {BM3D};
         \node[anchor=south] at (p16.north) {KSVD};
         
  \end{scope}
\end{tikzpicture}
}
\caption{Visual comparison of the spatial resolution on the CTP 528 high resolution module of the Catphan 600. Images are trained and tested on the full range DICOM image. Display window is [40, 400] HU. }
\label{fig:catphan}
\end{figure}
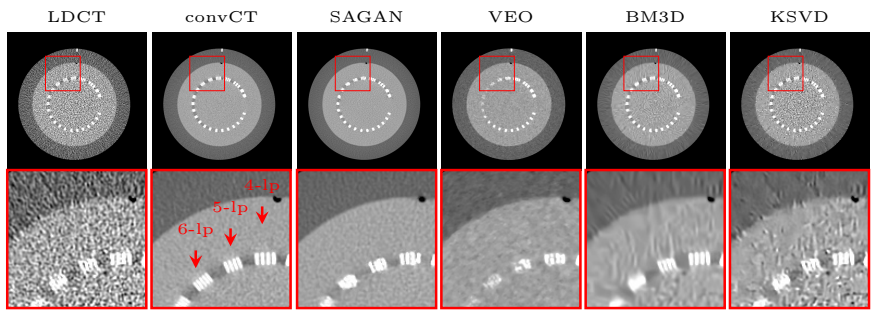

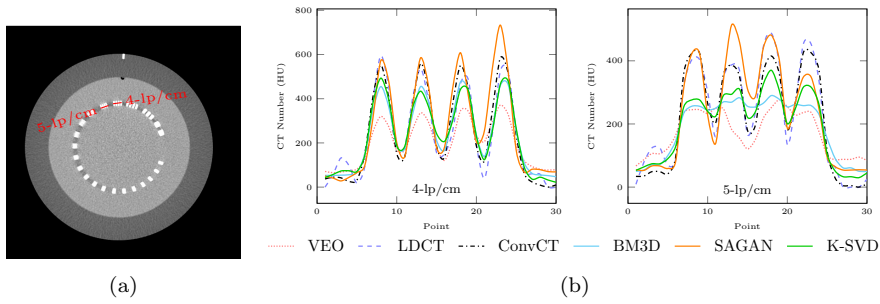
\begin{figure*}[tbp]
 \centering 
\pgfplotsset{every tick label/.append style={font=\tiny}}
\begin{subfigure}[b]{0.3\textwidth}
\centering
\begin{tikzpicture}
\begin{axis}[enlargelimits=false, axis on top, axis equal image,y dir=reverse,
	xtick=\empty,
	ytick=\empty,
        width =\textwidth,scale only axis
		]
\addplot graphics [xmin=0,xmax=51.2,ymin=0,ymax=51.2] {sr/target/0000148.jpg};
    \node[inner sep=0, outer sep =0] at (axis cs:20.3,18.2) (nodeA) {};
    \node[inner sep=0, outer sep =0] at (axis cs:22.6,17.4) (nodeB) {};
    \node[inner sep=0, outer sep =0] at (axis cs:23,17.15) (nodeC) {};
    \node[inner sep=0, outer sep =0] at (axis cs:25.5,16.8) (nodeD) {};
\draw[red,thin](nodeA)  -- (nodeB) node [midway, left, sloped] (TextNode1) {\tiny 5-lp/cm};
\draw[red,thin](nodeC)  -- (nodeD)node [midway, right, sloped] (TextNode2) {\tiny 4-lp/cm} ;

\end{axis}
\end{tikzpicture}
\caption*{(a)}
\end{subfigure}%
\begin{subfigure}[b]{0.7\textwidth}
\centering
\resizebox{0.48\textwidth}{!}{
\begin{tikzpicture}
\begin{axis}[xlabel=\tiny Point,ylabel=\tiny CT Number (HU),font = \small,xmin=0,xmax=30,width=0.8\textwidth,
		ylabel near ticks, xlabel near ticks, legend style={at={(1.1,0.5)},font=\tiny,anchor=west},mark size=1pt]
\addplot[ smooth,red!50,thick,densely dotted,mark options={solid}] plot coordinates {
(1,71)(2,65)(3,68)(4,74)(5,81)(6,127)(7,249)(8,320)(9,273)(10,167)(11,166)(12,261)(13,336)(14,293)(15,177)(16,121)(17,208)(18,345)(19,342)(20,233)(21,206)(22,297)(23,371)(24,327)(25,189)(26,119)(27,93)(28,81)(29,77)(30,79)
};
 \addplot[ smooth,blue!50,thick,dashed,mark options={solid}] plot coordinates {
(1,0)(2,57)(3,132)(4,98)(5,19)(6,95)(7,389)(8,595)(9,451)(10,193)(11,180)(12,403)(13,537)(14,388)(15,156)(16,150)(17,250)(18,468)(19,484)(20,193)(21,39)(22,240)(23,524)(24,522)(25,257)(26,74)(27,60)(28,58)(29,0)(30,0)
};
 \addplot[ smooth,black,thick,dashdotted,mark options={solid}] plot coordinates {
(1,37)(2,45)(3,42)(4,26)(5,31)(6,153)(7,406)(8,552)(9,403)(10,152)(11,137)(12,378)(13,560)(14,440)(15,185)(16,135)(17,371)(18,550)(19,440)(20,186)(21,118)(22,346)(23,584)(24,512)(25,236)(26,62)(27,23)(28,9)(29,0)(30,10)
};
 \addplot[ smooth,cyan!50,thick,mark options={solid}] plot coordinates {
(1,53)(2,53)(3,54)(4,56)(5,66)(6,151)(7,345)(8,455)(9,363)(10,187)(11,174)(12,330)(13,455)(14,385)(15,212)(16,166)(17,315)(18,476)(19,434)(20,235)(21,139)(22,270)(23,459)(24,462)(25,252)(26,110)(27,62)(28,55)(29,52)(30,47)
};
 \addplot[ smooth,orange,thick,mark options={solid}] plot coordinates {
(1,50)(2,41)(3,28)(4,44)(5,66)(6,99)(7,309)(8,572)(9,477)(10,201)(11,137)(12,370)(13,585)(14,455)(15,180)(16,181)(17,442)(18,608)(19,411)(20,204)(21,254)(22,536)(23,733)(24,518)(25,166)(26,68)(27,81)(28,79)(29,70)(30,69)
};
 \addplot[ smooth,green!80!black,thick,mark options={solid}] plot coordinates {
(1,43)(2,57)(3,73)(4,74)(5,71)(6,146)(7,378)(8,493)(9,395)(10,187)(11,184)(12,360)(13,434)(14,363)(15,242)(16,207)(17,298)(18,442)(19,436)(20,230)(21,124)(22,252)(23,465)(24,478)(25,265)(26,91)(27,36)(28,45)(29,32)(30,22)
};

 \node at (axis cs: 15, -10) {4-lp/cm};
\end{axis}
\end{tikzpicture}
}
\resizebox{0.48\textwidth}{!}{
\begin{tikzpicture}
\begin{axis}[xlabel=\tiny Point,ylabel=\tiny CT Number (HU),font = \small,xmin=0,xmax=30,width=0.8\textwidth,
		ylabel near ticks, xlabel near ticks, legend style={at={(1.1,0.5)},font=\tiny,anchor=west},mark size=1pt]
\addplot[ smooth,red!50,thick,densely dotted,mark options={solid}] plot coordinates {
(1,67)(2,87)(3,106)(4,106)(5,123)(6,165)(7,236)(8,245)(9,245)(10,233)(11,250)(12,247)(13,183)(14,155)(15,121)(16,151)(17,219)(18,252)(19,277)(20,255)(21,228)(22,236)(23,236)(24,174)(25,101)(26,87)(27,88)(28,88)(29,95)(30,84)
};
 \addplot[ smooth,blue!50,thick,dashed,mark options={solid}] plot coordinates {
(1,9)(2,79)(3,124)(4,124)(5,69)(6,90)(7,321)(8,403)(9,403)(10,302)(11,160)(12,348)(13,387)(14,369)(15,167)(16,240)(17,446)(18,487)(19,390)(20,135)(21,262)(22,452)(23,452)(24,332)(25,92)(26,63)(27,0)(28,0)(29,0)(30,25)
};
 \addplot[ smooth,black,thick,dashdotted,mark options={solid}] plot coordinates {
(1,34)(2,35)(3,49)(4,49)(5,42)(6,97)(7,360)(8,423)(9,423)(10,265)(11,203)(12,358)(13,386)(14,350)(15,180)(16,197)(17,366)(18,415)(19,358)(20,183)(21,267)(22,425)(23,425)(24,366)(25,114)(26,29)(27,5)(28,5)(29,0)(30,10)
};
 \addplot[ smooth,cyan!50,thick,mark options={solid}] plot coordinates {
(1,50)(2,54)(3,71)(4,71)(5,100)(6,147)(7,234)(8,256)(9,256)(10,244)(11,250)(12,270)(13,269)(14,283)(15,256)(16,254)(17,266)(18,290)(19,278)(20,254)(21,262)(22,258)(23,258)(24,229)(25,154)(26,91)(27,61)(28,61)(29,50)(30,50)
};
 \addplot[ smooth,orange,thick,mark options={solid}] plot coordinates {
(1,55)(2,53)(3,67)(4,67)(5,59)(6,83)(7,315)(8,422)(9,422)(10,228)(11,138)(12,343)(13,513)(14,459)(15,311)(16,287)(17,442)(18,481)(19,402)(20,187)(21,268)(22,348)(23,348)(24,245)(25,95)(26,65)(27,54)(28,54)(29,55)(30,54)
};
 \addplot[ smooth,green!80!black,thick,mark options={solid}] plot coordinates {
(1,53)(2,64)(3,74)(4,74)(5,90)(6,130)(7,254)(8,276)(9,276)(10,240)(11,224)(12,291)(13,294)(14,310)(15,221)(16,239)(17,327)(18,370)(19,306)(20,201)(21,248)(22,316)(23,316)(24,259)(25,105)(26,51)(27,33)(28,33)(29,30)(30,45)
};
 \node at (axis cs: 15, -10) {5-lp/cm};

\end{axis}

\end{tikzpicture}
}
\resizebox{\textwidth}{!}{
\begin{tikzpicture}
\begin{customlegend}[legend columns=7,legend style={align=left,draw=none,column sep=2ex,mark size=1pt,font=\large},legend entries={VEO, LDCT, ConvCT, BM3D, SAGAN, K-SVD},]
\addlegendimage{red!50,thick,densely dotted,mark options={solid}}
\addlegendimage{blue!50,thick,dashed,mark options={solid}}
\addlegendimage{black,thick,dashdotted,mark options={solid}}
\addlegendimage{cyan!50,thick,mark options={solid}}
\addlegendimage{orange,thick,mark options={solid}}
\addlegendimage{green!80!black,thick,mark options={solid}}
\end{customlegend}
\end{tikzpicture}
}
\caption*{(b)}
\end{subfigure}%

\caption{(a) shows the convCT of the CTP 528 high resolution module of the Catphan 600. (b) shows the line profile for the 4-line pair per centimeter  and 5-line pair per centimeter  of different reconstruction methods. 30 data points were sampled along the red line as marked in (a). }
\label{fig:lineprofile}
\end{figure*}

Figure~\ref{fig:catphan} gives the denoised visual result for the CTP 528 21 line pair high resolution module of the Catphan 600. The 4 and 5-line pairs is clearly distinguishable for LDCT and we can observe these line pairs equally well on SAGAN reconstructed images which suggests that the amount of spatial resolution loss is very small. 6-line pairs is distinguishable from the convCT but not the case for LDCT and all the reconstruction methods, which highlights the gap between the reconstruction methods and the convCT.         Figure~\ref{fig:lineprofile} shows the line profile along the  line drawn across the 4 and 5-line pairs. 30 points were sampled along the drawn line. SAGAN is the one among the comparative methods that achieves the highest spatial resolution. K-SVD behaves slightly better than BM3D and VEO demonstrates the lowest spatial resolution.

\subsection{Denoising results on the real piglet dataset}
\begin{figure*}
\newcommand{\ax}{0.5}
\newcommand{\ay}{0.7}
\newcommand{\bx}{0.3}
\newcommand{\by}{0.3}
\newcommand{\cx}{0.5}
\newcommand{\cy}{0.5}
\newcommand{\dx}{0.45}
\newcommand{\dy}{0.4}
\newcommand{\ex}{0.5}
\newcommand{\ey}{0.4}

\centering
\resizebox{\textwidth}{!}{
\begin{tikzpicture} [
    auto,
    line/.style     = { draw, thick, ->, shorten >=2pt,shorten <=2pt },
     every node/.append style={font=\tiny}
  ]
 \matrix [column sep=0.3mm, row sep=2mm,ampersand replacement=\&] {
 		 \node (p11)[inner sep=0] at (0,0){\magimage{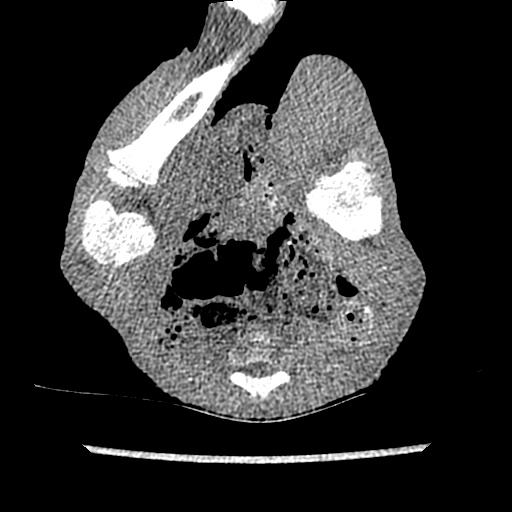}{0.13\textwidth}{\ax}{\ay}{0.13\textwidth}};     \&
		 \node (p12)[inner sep=0] at (0,0){\magimage{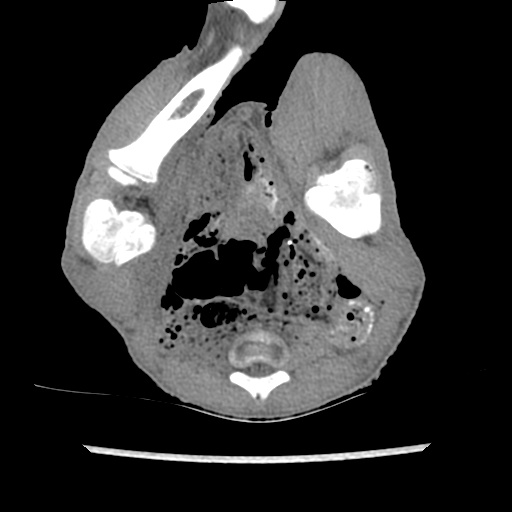}{0.13\textwidth}{\ax}{\ay}{0.13\textwidth}};      \&
 		 \node (p13)[inner sep=0] at (0,0){\magimage{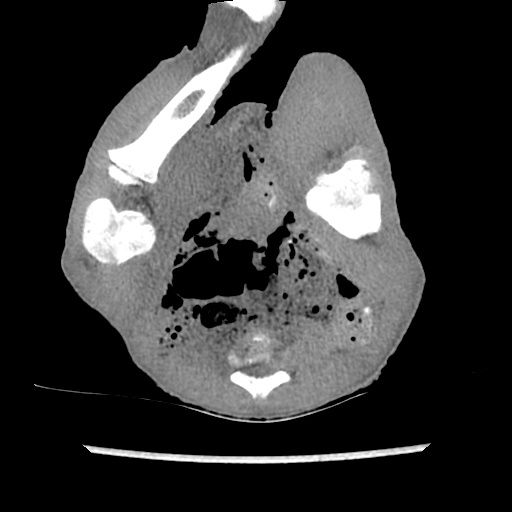}{0.13\textwidth}{\ax}{\ay}{0.13\textwidth}};     \&
 		 \node (p14)[inner sep=0] at (0,0){\magimage{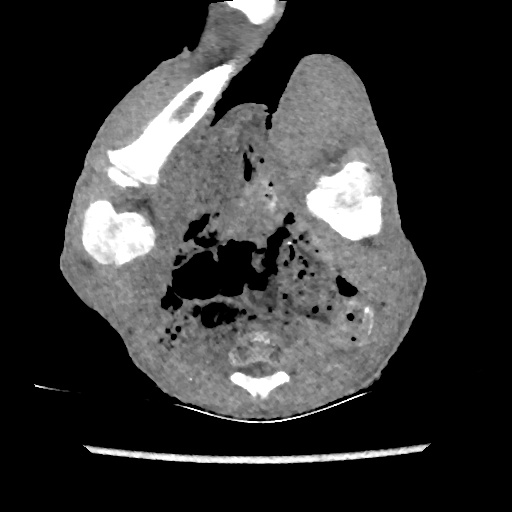}{0.13\textwidth}{\ax}{\ay}{0.13\textwidth}};     \&
 		 \node (p15)[inner sep=0] at (0,0){\magimage{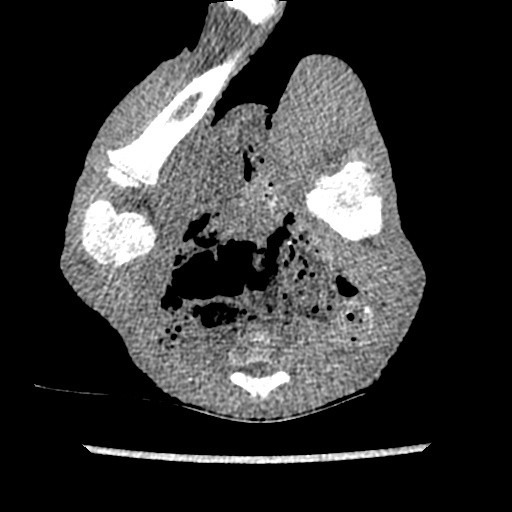}{0.13\textwidth}{\ax}{\ay}{0.13\textwidth}};     \&
 		 \node (p16)[inner sep=0] at (0,0){\magimage{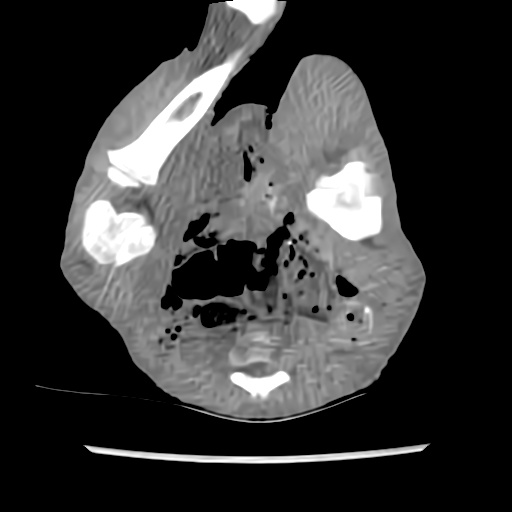}{0.13\textwidth}{\ax}{\ay}{0.13\textwidth}};     \&
 		 \node (p17)[inner sep=0] at (0,0){\magimage{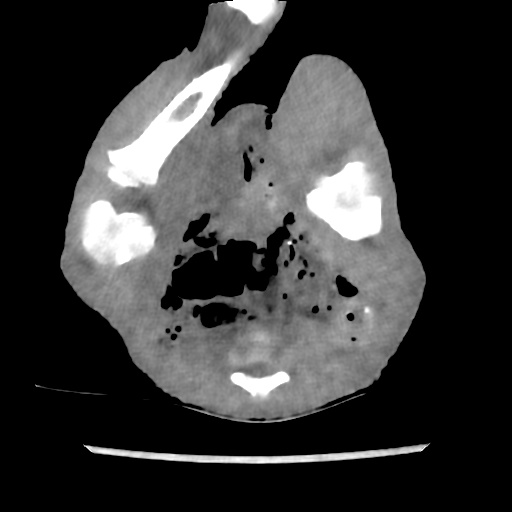}{0.13\textwidth}{\ax}{\ay}{0.13\textwidth}};     \&\\

		 \node (p21)[inner sep=0] at (0,0){\magimage{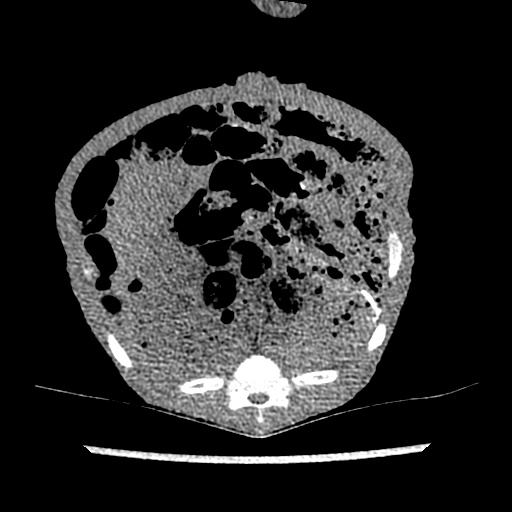}{0.13\textwidth}{\bx}{\by}{0.13\textwidth}};     \&
		 \node (p22)[inner sep=0] at (0,0){\magimage{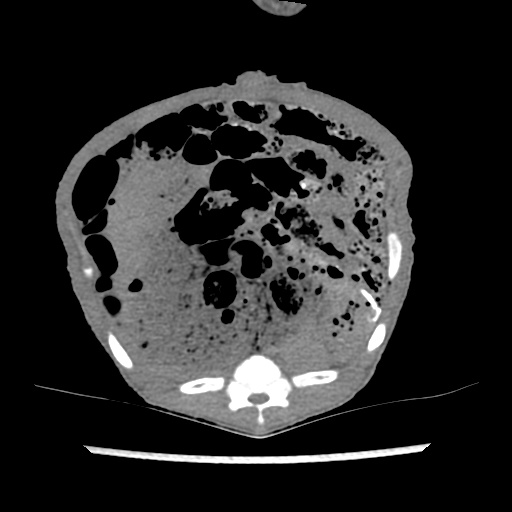}{0.13\textwidth}{\bx}{\by}{0.13\textwidth}};     \&
		 \node (p23)[inner sep=0] at (0,0){\magimage{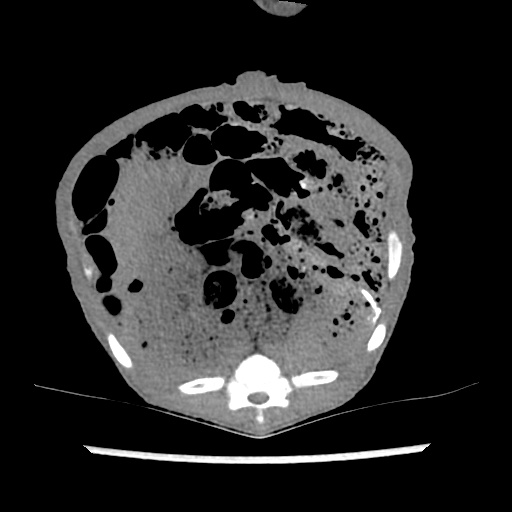}{0.13\textwidth}{\bx}{\by}{0.13\textwidth}};     \&
		 \node (p24)[inner sep=0] at (0,0){\magimage{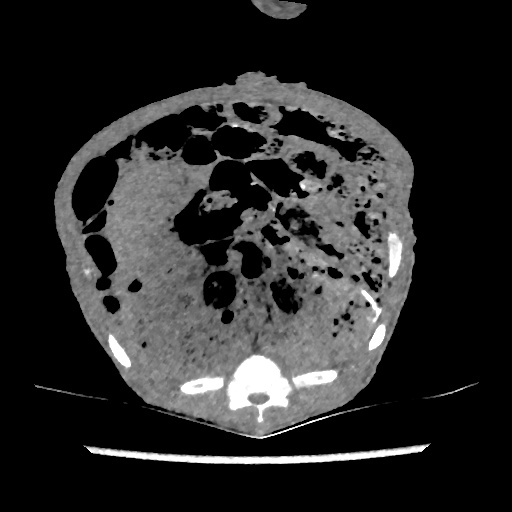}{0.13\textwidth}{\bx}{\by}{0.13\textwidth}};     \&
		 \node (p25)[inner sep=0] at (0,0){\magimage{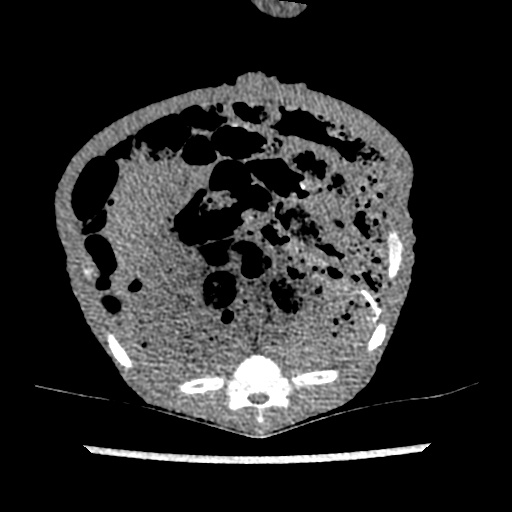}{0.13\textwidth}{\bx}{\by}{0.13\textwidth}};     \&
		 \node (p26)[inner sep=0] at (0,0){\magimage{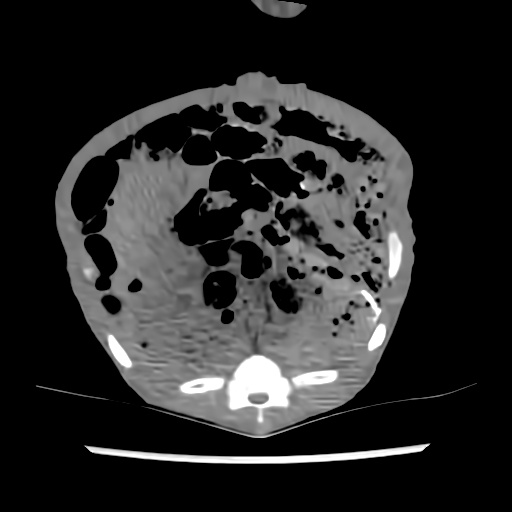}{0.13\textwidth}{\bx}{\by}{0.13\textwidth}};     \&
		 \node (p27)[inner sep=0] at (0,0){\magimage{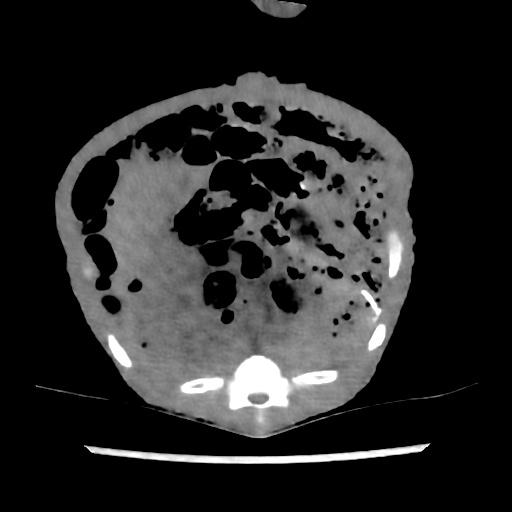}{0.13\textwidth}{\bx}{\by}{0.13\textwidth}};     \&\\

 		 \node (p31)[inner sep=0] at (0,0){\magimage{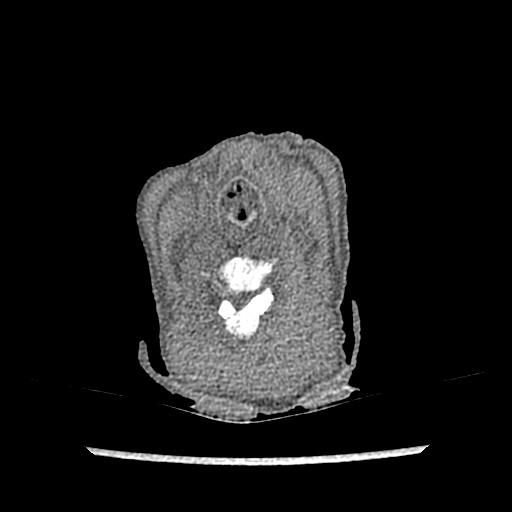}{0.13\textwidth}{\dx}{\dy}{0.13\textwidth}};     \&
 		 \node (p32)[inner sep=0] at (0,0){\magimage{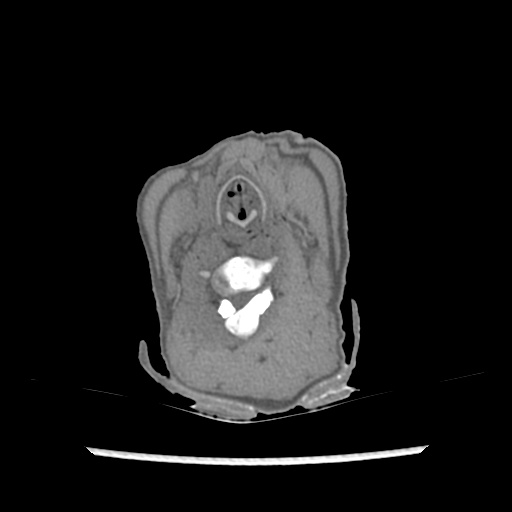}{0.13\textwidth}{\dx}{\dy}{0.13\textwidth}};     \&
 		 \node (p33)[inner sep=0] at (0,0){\magimage{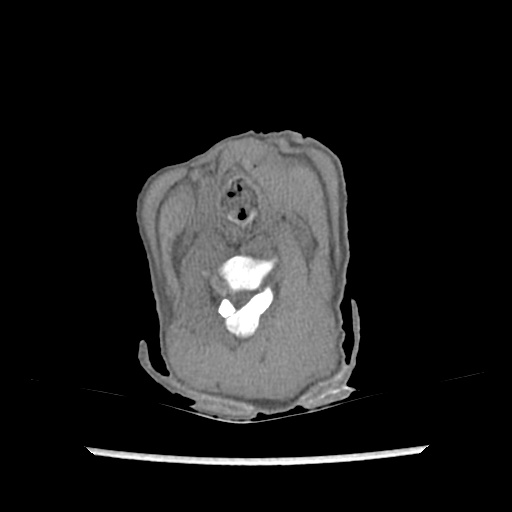}{0.13\textwidth}{\dx}{\dy}{0.13\textwidth}};     \&
 		 \node (p34)[inner sep=0] at (0,0){\magimage{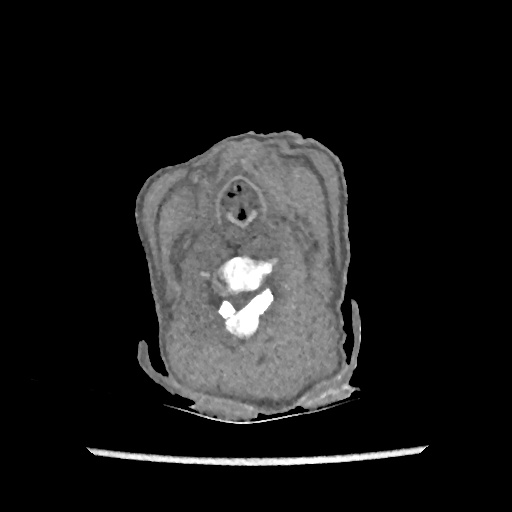}{0.13\textwidth}{\dx}{\dy}{0.13\textwidth}};     \&
 		 \node (p35)[inner sep=0] at (0,0){\magimage{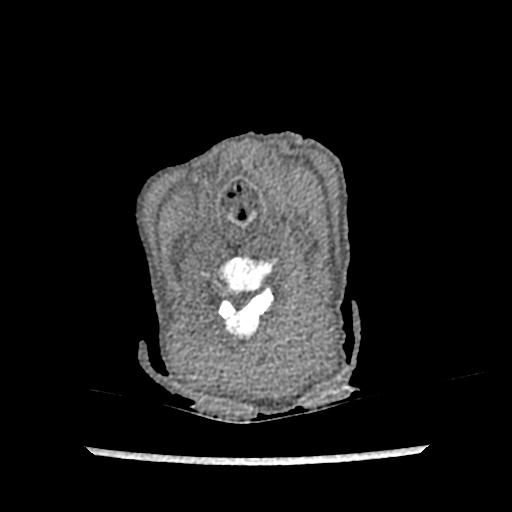}{0.13\textwidth}{\dx}{\dy}{0.13\textwidth}};     \&
		  \node (p36)[inner sep=0] at (0,0){\magimage{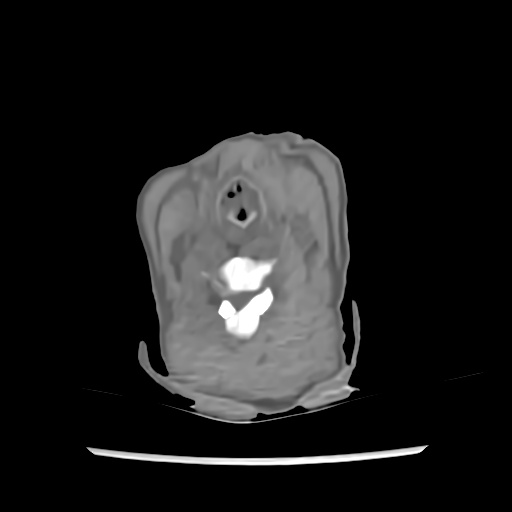}{0.13\textwidth}{\dx}{\dy}{0.13\textwidth}};     \&
 		 \node (p37)[inner sep=0] at (0,0){\magimage{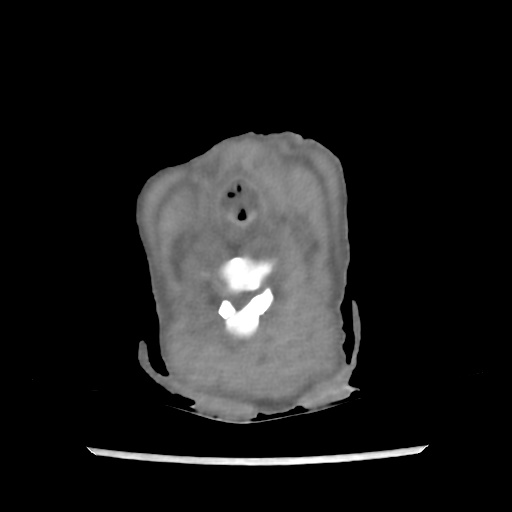}{0.13\textwidth}{\dx}{\dy}{0.13\textwidth}};     \&\\
                       };
  \begin{scope} [every path/.style=line]
     \node[anchor=south] at (p11.north) {LDCT};
     \node[anchor=south] at (p12.north) {convCT};
     \node[anchor=south] at (p13.north) {SAGAN};
     \node[anchor=south] at (p14.north) {VEO};
     \node[anchor=south] at (p15.north) {ASIR};
     \node[anchor=south] at (p16.north) {BM3D};
     \node[anchor=south] at (p17.north) {K-SVD};
  \end{scope}
\end{tikzpicture}
}
\caption{Visual examples for the denoised images on the piglet dataset. Rows are  3 selected samples from pelvis to head each with a zoomed up local region. The first column is the LDCT (5\% of full dose reconstructed by FBP, 0.71 mSv). The second column is the convCT (100\% dose reconstructed by FBP, 14.14 mSv). The last 5 columns are results from different denoising methods. Display window is [40, 400].}
\label{fig:body}
\end{figure*}
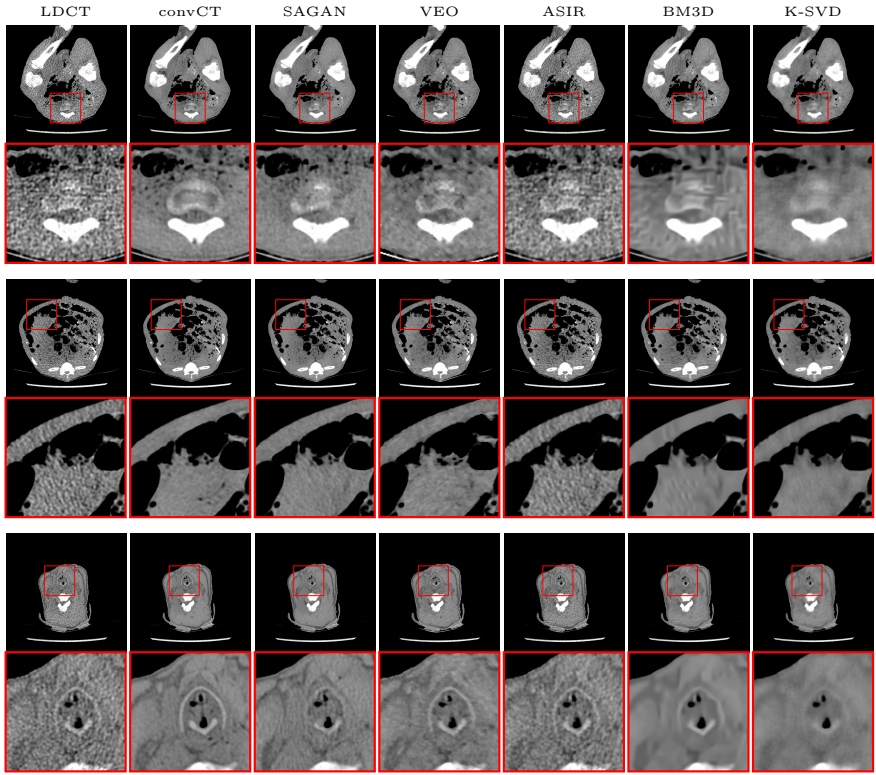

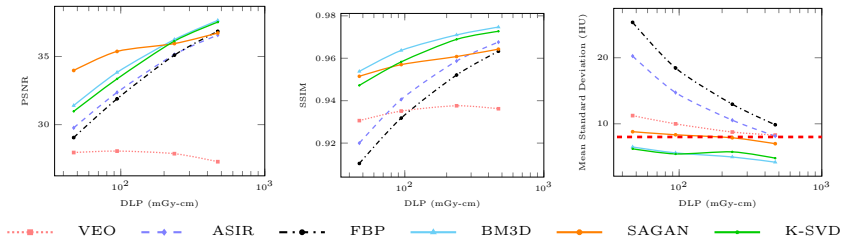
\begin{figure*}[htbp]
 \centering 
\pgfplotsset{every tick label/.append style={font=\tiny}}
\resizebox {0.3\textwidth} {!} {
\begin{tikzpicture}
\begin{axis}[xlabel=\tiny DLP (mGy-cm),ylabel=\tiny PSNR,xmin=0,xmax=1000,width=0.5\textwidth,
ylabel near ticks,
		 xlabel near ticks, legend style={at={(1.1,0.5)},anchor=west},xmode = log,mark size=0.7pt]
\addplot[ smooth,red!50,thick,mark=square*,densely dotted,mark options={solid}] plot coordinates {
(47.16, 27.9517)(94.32, 28.0576)(235.81, 27.8665)(471.62, 27.2778)
};
 \addplot[ smooth,blue!50,thick,mark=diamond*,dashed,mark options={solid}] plot coordinates {
(47.16, 29.7621)(94.32, 32.3641)(235.81, 35.1410)(471.62, 36.5728)
};
 \addplot[ smooth,black,thick,mark=otimes*,dashdotted,mark options={solid}] plot coordinates {
(47.16, 29.0489)(94.32, 31.8943)(235.81, 35.0985)(471.62, 36.8473)
};
 \addplot[ smooth,cyan!50,thick,mark=triangle*,mark options={solid}] plot coordinates {
(47.16, 31.4006)(94.32, 33.8336)(235.81, 36.2443)(471.62, 37.6448)
};
 \addplot[ smooth,orange,thick,mark=*,mark options={solid}] plot coordinates {
(47.16, 33.9736)(94.32, 35.3719)(235.81, 35.9469)(471.62, 36.7240)
};
 \addplot[ smooth,green!80!black,thick,mark=+,mark options={solid}] plot coordinates {
(47.16, 30.9734)(94.32, 33.3574)(235.81, 36.1295)(471.62, 37.5226)
};
\end{axis}
\end{tikzpicture}
}
\resizebox {0.3\textwidth} {!} {
\begin{tikzpicture}
\begin{axis}[xlabel=\tiny DLP (mGy-cm),ylabel=\tiny SSIM,xmin=0,xmax=1000,width=0.5\textwidth,
ylabel near ticks, xlabel near ticks, legend style={at={(1.1,0.5)},anchor=west},xmode = log,mark size=0.7pt]
\addplot[ smooth,red!50,thick,mark=square*,densely dotted,mark options={solid}] plot coordinates {
(47.16, 0.9306)(94.32, 0.9351)(235.81, 0.9376)(471.62, 0.9362)
};
 \addplot[ smooth,blue!50,thick,mark=diamond*,dashed,mark options={solid}] plot coordinates {
(47.16, 0.9200)(94.32, 0.9406)(235.81, 0.9587)(471.62, 0.9676)
};
 \addplot[ smooth,black,thick,mark=otimes*,dashdotted,mark options={solid}] plot coordinates {
(47.16, 0.9104)(94.32, 0.9318)(235.81, 0.9521)(471.62, 0.9633)
};
 \addplot[ smooth,cyan!50,thick,mark=triangle*,mark options={solid}] plot coordinates {
(47.16, 0.9538)(94.32, 0.9637)(235.81, 0.9710)(471.62, 0.9747)
};
 \addplot[ smooth,orange,thick,mark=*,mark options={solid}] plot coordinates {
(47.16, 0.9515)(94.32, 0.9570)(235.81, 0.9608)(471.62, 0.9643)
};
 \addplot[ smooth,green!80!black,thick,mark=+,mark options={solid}] plot coordinates {
(47.16, 0.9472)(94.32, 0.9583)(235.81, 0.9689)(471.62, 0.9727)
};
\end{axis}
\end{tikzpicture}
}
\resizebox {0.3\textwidth} {!} {
\begin{tikzpicture}
\begin{axis}[xlabel=\tiny DLP (mGy-cm),ylabel=\tiny Mean Standard Deviation (HU),xmin=0,xmax=1000,width=0.5\textwidth,
ylabel near ticks, xlabel near ticks, legend style={at={(1.1,0.5)},anchor=west},xmode = log,mark size=0.7pt,legend pos=outer north east]
\addplot[ smooth,red!50,thick,mark=square*,densely dotted,mark options={solid}] plot coordinates {
(47.16, 11.2393)(94.32, 9.9866)(235.81, 8.7512)(471.62, 8.2946)
};
 \addplot[ smooth,blue!50,thick,mark=diamond*,dashed,mark options={solid}] plot coordinates {
(47.16, 20.2519)(94.32, 14.7282)(235.81, 10.5456)(471.62, 8.1407)
};
 \addplot[ smooth,black,thick,mark=otimes*,dashdotted,mark options={solid}] plot coordinates {
(47.16, 25.3548)(94.32, 18.4441)(235.81, 12.9640)(471.62, 9.8561)
};
 \addplot[ smooth,cyan!50,thick,mark=triangle*,mark options={solid}] plot coordinates {
(47.16, 6.4864)(94.32, 5.5914)(235.81, 4.9763)(471.62, 4.1834)
};
 \addplot[ smooth,orange,thick,mark=*,mark options={solid}] plot coordinates {
(47.16, 8.8021)(94.32, 8.3352)(235.81, 7.8834)(471.62, 6.9869)
};
 \addplot[ smooth,green!80!black,thick,mark=+,mark options={solid}] plot coordinates {
(47.16, 6.2140)(94.32, 5.4367)(235.81, 5.7486)(471.62, 4.7949)
};

 \draw [ultra thick, dashed, draw=red] 
        (axis cs: 0,8.0064) -- (axis cs: 1000, 8.0064)
        node[pos=0.5, above] {};
\end{axis}
\end{tikzpicture}
}
\begin{tikzpicture}
\begin{customlegend}[legend columns=7,legend style={align=left,draw=none,column sep=2ex,mark size=0.7pt,font=\tiny},legend entries={VEO, ASIR, FBP, BM3D, SAGAN, K-SVD}]
\addlegendimage{red!50,thick,mark=square*,densely dotted,mark options={solid}}
\addlegendimage{blue!50,thick,mark=diamond*,dashed,mark options={solid}}
\addlegendimage{black,thick,mark=otimes*,dashdotted,mark options={solid}}
\addlegendimage{cyan!50,thick,mark=triangle*,mark options={solid}}
\addlegendimage{orange,thick,mark=*,mark options={solid}}
\addlegendimage{green!80!black,thick,mark=+,mark options={solid}}
\end{customlegend}
\end{tikzpicture}

\caption{The left two figures show the PSNR and SSIM  plotted against dose-length product (DLP) for different reconstruction methods for the piglet dataset. The rightmost figure gives the mean standard deviation of image noise against dose-length product (DLP) for different reconstruction methods. Red dashed line refers to the standard deviation of the FBP reconstructed convCT. }
\label{fig:pigpsnr}
\end{figure*}

Here we plotted a line graph of the PSNR and SSIM against the dosage in Figure~\ref{fig:pigpsnr} for all the comparator methods. It can be seen that all methods except VEO have their performance improved with the increase of dose in terms of PSNR. SSIM is less affected because it penalizes structural differences rather than the pixel-wise difference. The average SSIM measure in the lowest dose level for SAGAN is 0.95 which is slightly higher than that of the second highest dose level  for FBP. Figure~\ref{fig:body} shows some visual examples from different anatomic region (from head to pelvis) at the lowest dose level and their reconstruction by all the comparator methods. We can see clearly that SAGAN produces results that are more visually appealing than the others.

Figure~\ref{fig:pigpsnr} also shows the mean standard deviation of CT numbers on 42 hand selected rectangular homogeneous regions as a direct measure of noise level. The red horizontal dashed line is the performance of the convCT and serves as reference and it can be seen that the available commerical methods do not surpass the reference line. In general, the mean standard deviation of SAGAN results  are pretty constant across all dose levels and  very close to the convCT. It implicitly shows that SAGAN can simulate the statistical properties of convCT. BM3D and K-SVD on the other hand obtained smaller numbers by over-smoothing the result images.   At the highest noise level, the measure was 25.35 for FBP and 8.80 for SAGAN,  corresponding to a noise reduction factor of 2.88.   Considering both the quantitative measures and the visual appearance, SAGAN is no doubt the best method among the comparators in the highest noise level.

\subsection{Denoising results on the clinical patient data with unknown dose level \protect\footnote{Testing code is available at https://github.com/xinario/SAGAN}}
\begin{figure}[!t]
\centering
\begin{tikzpicture} [
    auto,
    line/.style     = { draw, thick, ->, shorten >=2pt,shorten <=2pt },    every node/.append style={font=\tiny}
  ]
\newcommand{\ax}{0.55}
\newcommand{\ay}{0.55}
\newcommand{\imgnum}{0000184}
 \matrix [column sep=0.3mm, row sep=5mm,ampersand replacement=\&] {
 		 \node (p11)[inner sep=0] at (0,0){\magimage{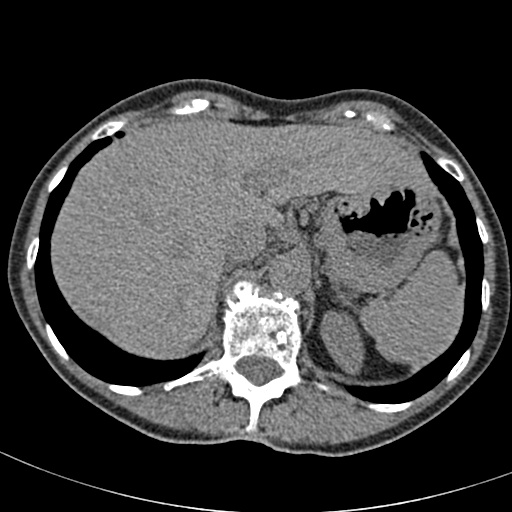}{0.24\textwidth}{0.5}{0.35}{0.24\textwidth}};     \&
		 \node (p12)[inner sep=0] at (0,0){\magimageblue{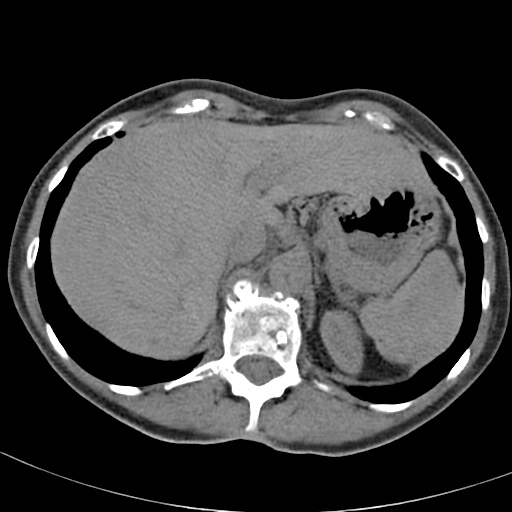}{0.24\textwidth}{0.5}{0.35}{0.24\textwidth}};     \&
		 \node (p13)[inner sep=0] at (0,0){\magimage{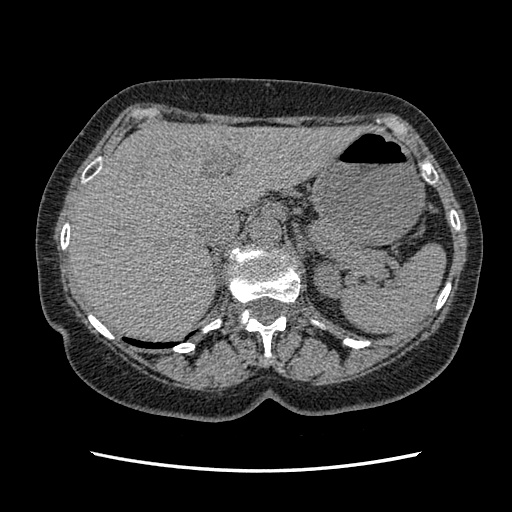}{0.24\textwidth}{0.5}{0.7}{0.24\textwidth}};     \&
		 \node (p14)[inner sep=0] at (0,0){\magimageblue{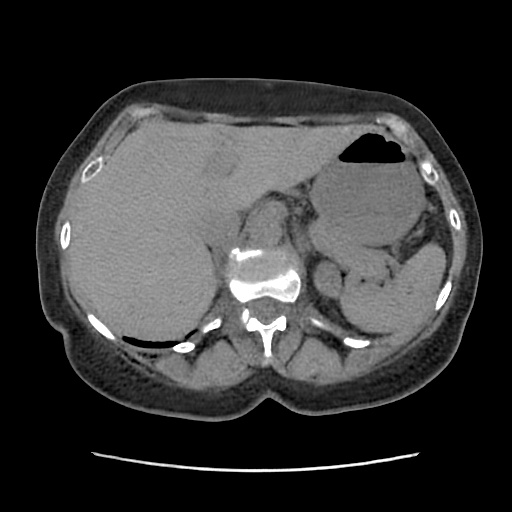}{0.24\textwidth}{0.5}{0.7}{0.24\textwidth}};     \&\\
                       };
  \begin{scope} [every path/.style=line]
      \node[anchor=south] at (p11.north) {LDCT ($17.06$)};
       \node[anchor=south] at (p12.north) {SAGAN ($8.72$)};
       \node[anchor=south] at (p13.north) {LDCT ($22.98$)};
       \node[anchor=south] at (p14.north) {SAGAN ($9.83$)};
       
      \node[anchor=north,xshift=0.065\textwidth,yshift=-10pt] at (p11.south) {(a)};
       \node[anchor=north,xshift=0.065\textwidth,yshift=-10pt] at (p13.south) {(b)};


  \end{scope}
\end{tikzpicture}

\caption{SAGAN denoising results on the clinical patient data (from the Kaggle Data Science Bowl 2017) with unknown dose levels. Image pairs of (a), (b) come from 2 different patients.  The noise level of the liver in (a) and lung nodule in (b) were shown below the images. Display window of  is [40, 400] HU for (a) and [-700, 1500] HU for (b). Numbers in the braces  are the noise level computed from 20 homogeneous regions selected from the patient scan.}
\label{fig:patient}
\end{figure}
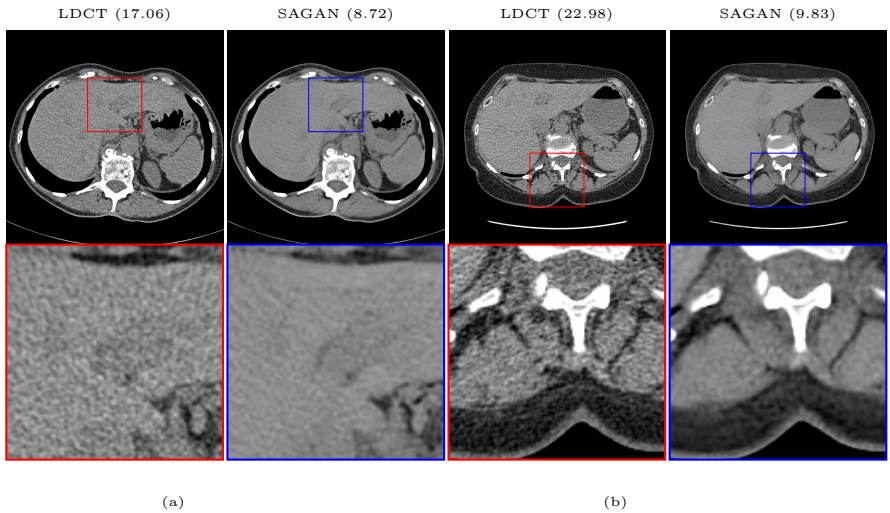

Figure~\ref{fig:patient} shows the results on the clinical patient data with unknown dose levels. The dose level of these data are unlikely to coincide with the dose level of our training set but we can see that it performs reasonable well on these images with increased CNR.

\section{Discussion}\label{discussion}
These quantitative results demonstrate that SAGAN excels in recovering underlying structures in great uncertainty. The adversarially trained discriminator guarantees the denoised texture to be close to convCT.  This is an advantage over VEO, which produces a different texture.  The sharpness detection network guarantees that the generated CT is with similar sharpness as the convCT. Another advantage is time efficiency. Neural network based methods, including SAGAN only need one forward pass in the testing and the task could be accomplished in less than a second. BM3D showed better denoising at the highest dose, however SAGAN was better at lower doses. BM3D and K-SVD also had evident streak artifacts across the image surface at low doses as seen in Figure 3.

Another phenomenon we have observed is that the SAGAN can also help to mitigate the streak artifacts. As can be seen from the first row of Figure~\ref{fig:edgy}, the lower half of the convCT had  mild streak artifacts but was less evident in the SAGAN result (4th column). The reason we think is that the discriminator discriminates patches and the number of patches containing artifacts is significantly smaller than the number of normal ones. Therefore these patches were considered as outliers in the discrimination process. A straightforward extension of this work would be for limited view CT reconstruction.

The proposed generator here adopts the Unet\cite{ronneberger2015u,  isola2016image} architecture and incorporates the residual connection for the ease of training. We have empirically demonstrate its effectiveness on the denoising task and have observed much more stable training statistics than the comparators in the adversarial training scheme. 

The sharpness-aware loss proposed here is similar to the methodology of the content loss as used in~\cite{johnson2016perceptual,ledig2016photo} but differs in the final purpose. The similarity lies in that we both measure the high-level features of the generated and input image. In their work,  the high-level features are from the middle layer of the pre-trained VGG network~\cite{simonyan2014very}  and used to ensure the perceptual similarity. On the contrary, the high-level features used here are  extracted from a specifically trained network and directly correspond to the visual sharpness.

The work of Wolterink et al.~\cite{wolterink2017generative} and Yang et al.~\cite{yang2017low} also employed GAN for  CT denoising and some technical differences from ours have been highlighted in  section~\ref{related}. Here we also want to emphasize two of their weaknesses.  These two works  either centered on cardiac CT or abdominal CT. It is not clear whether their trained model can be applied to CTs of different anatomies. Our work considered a wider range of anatomic regions ranging from head to pelvis and has demonstrated that a single network in cGAN setting would be sufficient to denoise CT of the whole body. Moreover, their work only employed a single dose level in the training whereas ours  covered a wider range of dose levels. We have also empirically shown that the trained model not only suitable for denoising images with the training dose level but also applicable to unseen dose levels as long as the noise level is within our training range.

The dose reduction achieved by SAGAN is very high. According to the measurement of PSNR and SSIM, SAGAN reconstructed result in the lowest dose level has a measurement almost equivalent to the CT in the second highest dose level  (7.07 mSv), corresponding to a dose reduction of 90\%. Meanwhile if we measure the dose reduction factor with respect to the mean standard deviation of attenuation, SAGAN's result in the lowest dose level (0.71 mSv) has a noise level similar to that of the convCT (14.14 mSv), corresponding to a dose reduction of 95\%. Each measurement has their own strengths and weaknesses in measuring the CT image quality. PSNR and SSIM take into consideration  the spatial information but would penalize a lot of the texture difference even with similar underlying image content. For example, although VEO has been shown to have superior performance in many clinical studies than ASIR and FBP and has received clearance released by the Food and Drug Administration of the USA, it obtained the worst PSNR and SSIM measurements. On the other hand, mean standard deviation of attenuation direct measures  the noise level, but completely discarded the spatial information.  Therefore, we reported both results for the sake of fair comparison. The work of Suzuki et al.~\cite{suzuki2017neural} reported a dose reduction of 90\% from 1.1 mSv to 0.11 mSv with MTANN. Their network is patched based and need to train multiple networks corresponding to different anatomic regions.  It is unclear how much dose reduction was achieved for the other deep learning based approaches~\cite{yang2017ct, kang2016deep, chen2017low, wolterink2017generative}.

We evaluated the spatial resolution of the reconstructed image using the Catphan 600 high resolution module. This analysis was generally missed for the other deep learning based approaches. These methods could achieve high PSNR and SSIM by directly treat them as the optimization objective  but at the cost of losing spatial resolution. GAN based methods, including ours mitigate this problem by incorporating the adversarial objective. We think it is crucial to bring the spatial resolution  into assessment for the deep learning based  approaches when PSNR and SSIM become less effective.  An alternative way of quantifying the denoising performance would be to measure the performance of subsequent higher level tasks, e.g. lung nodule detection, anatomical region segmentation. We would like to leave this to the future work.

\section{Conclusion}\label{conclusion}

In this paper, we have proposed  sharpness aware network for low dose CT denoising. It utilizes both the adversarial loss and the sharpness loss to leverage the blur effect faced by image based denoising method, especially under high noise levels. The proposed SAGAN achieves improved performance in the quantitative assessment and the visual results are more appealing than the tested competitors.

However, we acknowledge that there are some limitations of this work that are waiting to be solved in the future.  First of all, the sharpness detection network is trained to compute the sharpness metric of~\cite{yi2016lbp} which is not very sensitive to just noticeable blur. This could limit the final sharpness of the denoised image, especially some small low contrast regions.

Second, for all the deep learning based methods, the network need to be trained against a specific dosage level. Even though we  trained our method on a wild range of doses and have applied it to clinical patient data, the analysis is mostly centred on the visual quality assessment of the denoised image.  The image diagnosis performance  in clinical practice remains to be evaluated.

\begin{acknowledgements}
The authors would like to thank Troy Anderson for the acquisition of the piglet dataset.
\end{acknowledgements}


\bibliographystyle{spbasic} 

\bibliography{ct_denoise,ldct_simulation,gan,common,blurseg}

\end{document}